\documentclass[10pt,twocolumn,letterpaper]{article}

\usepackage{iccv}
\usepackage{times}
\usepackage{epsfig}
\usepackage{graphicx}
\usepackage{amsmath}
\usepackage{amssymb}
\usepackage{bm}
\usepackage{nicefrac}
\usepackage{booktabs}
\usepackage{multirow}
\usepackage{multicol}
\usepackage{makecell}
\usepackage{array}
\usepackage{enumitem}
\usepackage{gensymb}  %
\usepackage{microtype}  %
\usepackage[percent]{overpic} %
\usepackage{rotating}
\usepackage{adjustbox}
\usepackage{siunitx}
\usepackage{pifont}
\usepackage{stmaryrd}  %
\usepackage[format=plain,labelformat=simple,labelsep=period,font=small,compatibility=false]{caption}
\usepackage[dvipsnames]{xcolor}
\usepackage{svg}

\usepackage[pagebackref=true,breaklinks=true,letterpaper=true,colorlinks,bookmarks=false]{hyperref}
\hypersetup{ %
    citecolor=ForestGreen,
    urlcolor=MidnightBlue,
}

\newcommand{\PAR}[1]{\vskip4pt \noindent{\bf #1~}}
\renewcommand{\*}[1]{\bm{\mathrm{#1}}}

\renewcommand{\b}[1]{\textbf{#1}}
\newcommand{\red}[1]{\textcolor{red}{#1}}
\newcommand{\green}[1]{\textcolor{green}{#1}}

\newcommand{\0}{\phantom{0}}

\newcommand{\supp}{Appendix} %
\newif\ifproceedings  %
\proceedingstrue
\proceedingsfalse
\newif\ifsupponly
\supponlyfalse

\DeclareFontFamily{U}{mathb}{}
\DeclareFontShape{U}{mathb}{m}{n}{
  <-5.5> mathb5
  <5.5-6.5> mathb6
  <6.5-7.5> mathb7
  <7.5-8.5> mathb8
  <8.5-9.5> mathb9
  <9.5-11.5> mathb10
  <11.5-> mathb12
}{}
\DeclareSymbolFont{mathb}{U}{mathb}{m}{n}
\DeclareMathSymbol{\drsh}{3}{mathb}{"EB}
\newcommand{\cornerarrow}{\raisebox{1pt}{$\drsh\ $}}

\usepackage[capitalize]{cleveref}
\crefname{section}{Sec.}{Secs.}
\Crefname{section}{Section}{Sections}
\Crefname{table}{Table}{Tables}
\crefname{table}{Tab.}{Tabs.}

\iccvfinalcopy %
\def\iccvPaperID{4434} %

\ificcvfinal
\ifproceedings
\fi
\fi

\begin{document}

\title{LightGlue: Local Feature Matching at Light Speed}

\author{%
Philipp Lindenberger$^{1}$\hspace{.2in}
Paul-Edouard Sarlin$^{1}$\hspace{.2in}
Marc Pollefeys$^{1,2}$
\vspace{.08in}\\
$^{1}$ ETH Zurich\hspace{0.2in}
$^{2}$ Microsoft Mixed Reality \& AI Lab
}

\maketitle
\ificcvfinal
\ifproceedings
\thispagestyle{empty}
\fi
\fi

\begin{abstract}
We introduce LightGlue, a deep neural network that learns to match local features across images.
We revisit multiple design decisions of SuperGlue, the state of the art in sparse matching, and derive simple but effective improvements.
Cumulatively, they make LightGlue more efficient --~in terms of both memory and computation, more accurate, and much easier to train.
One key property is that LightGlue is adaptive to the difficulty of the problem: the inference is much faster on image pairs that are intuitively easy to match, for example because of a larger visual overlap or limited appearance change.
This opens up exciting prospects for deploying deep matchers in latency-sensitive applications like 3D reconstruction.
The code and trained models are publicly available at \href{https://github.com/cvg/LightGlue}{\texttt{github.com/cvg/LightGlue}}.
\end{abstract}

\section{Introduction}
Finding correspondences between two images is a fundamental building block of many computer vision applications like camera tracking and 3D mapping.
The most common approach to image matching relies on sparse interest points that are matched using high-dimensional representations encoding their local visual appearance. %
Reliably describing each point is challenging in conditions that exhibit symmetries, weak texture, or appearance changes due to varying viewpoint and lighting.
To reject outliers that arise from occlusion and missing points, such representations should also be discriminative.
This yields two conflicting objectives, robustness and uniqueness, that are hard to satisfy.

To address these limitations, SuperGlue~\cite{sarlin2020superglue} introduced a new paradigm -- a deep network that considers both images at the same time to jointly match sparse points and reject outliers.
It leverages the powerful Transformer model~\cite{vaswani2017attention} to learn to match challenging image pairs from large datasets.
This yields robust image matching in both indoor and outdoor environments.
SuperGlue is highly effective for visual localization in challenging conditions~\cite{sattler2018benchmarking,sarlin2019coarse,sarlin21pixloc,sarlin2022lamar}
and generalizes well to other tasks like aerial matching~\cite{ZHANG2021176}, object pose estimation~\cite{sun2022onepose}, and even fish re-identification~\cite{pedersen2022re}.

These improvements are however computationally expensive, while the efficiency of image matching is critical for tasks that require a low latency, like tracking, or a high processing volume, like large-scale mapping.
Additionally, SuperGlue, as with other Transformer-based models, is notoriously hard to train, requiring computing resources that are inaccessible to many practitioners.
Follow-up works~\cite{sgmnet,clustergnn} have thus failed to reach the performance of the original SuperGlue model.
Yet, since its initial publication, Transformers have been extensively studied, improved, and applied to numerous language~\cite{bert,gpt,lamda} and vision~\cite{vit,dino,percieverio} tasks.

\begin{figure}[t]
    \centering
    \includegraphics[width=\linewidth]{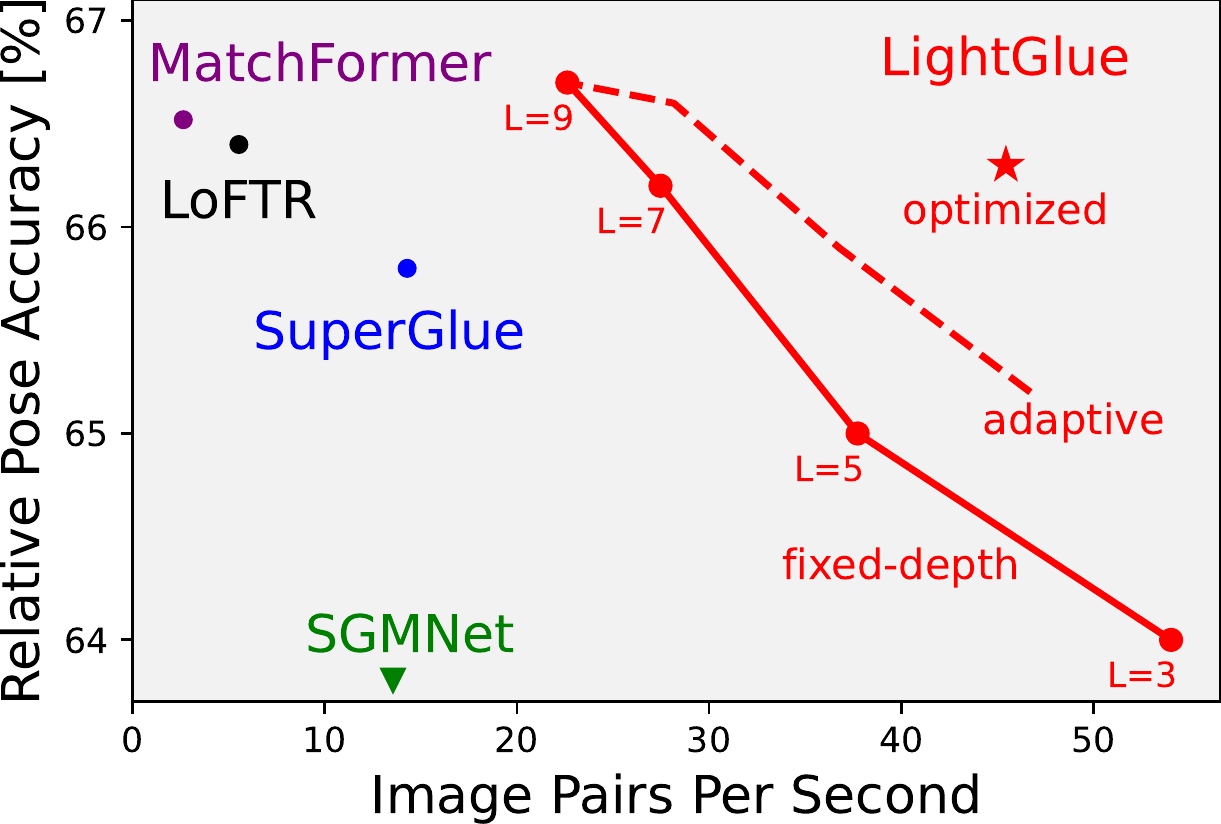}%
    \vspace{1mm}
    
    \caption{\textbf{LightGlue matches sparse features faster and better}
    than existing approaches like SuperGlue.
    Its adaptive stopping mechanism gives a fine-grained control over the speed vs.\ accuracy trade-off.
    Our final, optimized model {\large\red{$\star$}} delivers an accuracy closer to the dense matcher LoFTR at an 8$\times$ higher speed, here in typical outdoor conditions.
    }%
    \label{fig:teaser}%
\end{figure}

In this paper, we draw on these insights to design LightGlue, a deep network that is more accurate, more efficient, and easier to train than SuperGlue.
We revisit its design decisions and combine numerous simple, yet effective, architecture modifications.
We distill a recipe to train high-performance deep matchers with limited resources, reaching state-of-the-art accuracy within just a few GPU-days.
As shown in Figure~\ref{fig:teaser}, LightGlue is Pareto-optimal on the efficiency-accuracy trade-off when compared to existing sparse and dense matchers.

\begin{figure}[t]
    \centering
    \includegraphics[width=\linewidth]{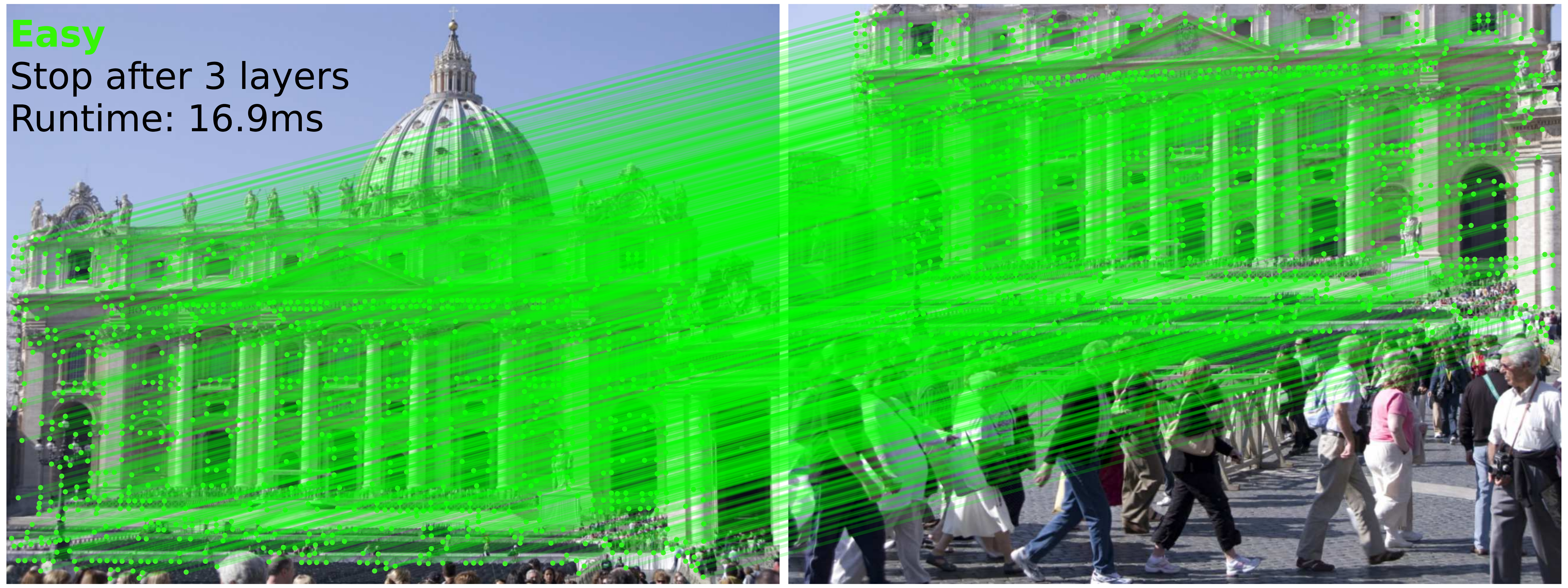}
    \includegraphics[width=\linewidth]{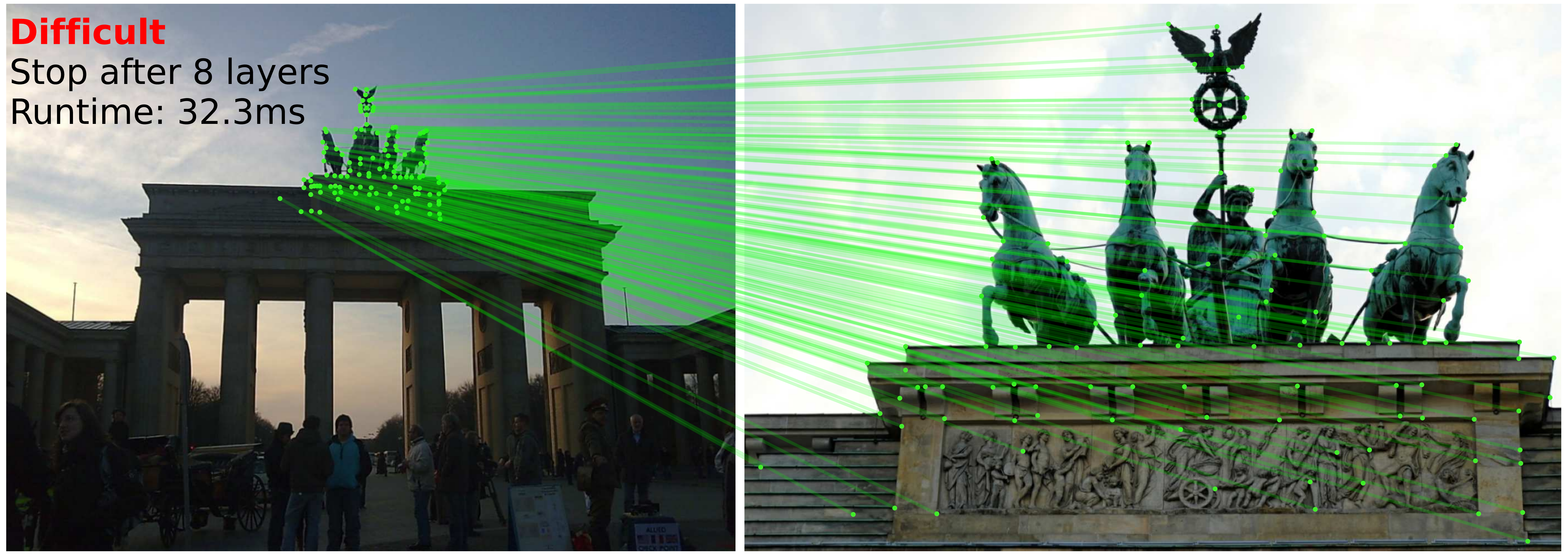}
    \caption{\textbf{Depth adaptivity.}
    LigthGlue is faster at matching easy image pairs (top) than difficult ones (bottom) because it can stop at earlier layers when its predictions are confident.
    }%
    \label{fig:demo}%
\end{figure}

Unlike previous approaches, LightGlue is adaptive to the difficulty of each image pair, which varies based on the amount of visual overlap, appearance changes, or discriminative information.
Figure~\ref{fig:demo} shows that the inference is thus much faster on pairs that are intuitively easy to match than on challenging ones, a behavior that is reminiscent of how humans process visual information.
This is achieved by 1) predicting a set of correspondences after each computational blocks, and 2) enabling the model to introspect them and predict whether further computation is required.
LigthGlue also discards at an early stage points that are not matchable, thus focusing its attention on the covisible area.

Our experiments show that LightGlue is a plug-and-play replacement to SuperGlue: it predicts strong matches from two sets of local features, at a fraction of the run time.
This opens up exciting prospects for deploying deep matchers in latency-sensitive applications like SLAM~\cite{orbslam,cadena2016past}
or reconstructing larger scenes from crowd-sourced data~\cite{heinly2015reconstructing,schoenberger2016sfm,lindenberger2021pixsfm,sarlin2022lamar}.
The LightGlue model and its training code will be released publicly with a permissive license.

\begin{figure*}[t]
    \centering
    \includegraphics[width=\linewidth]{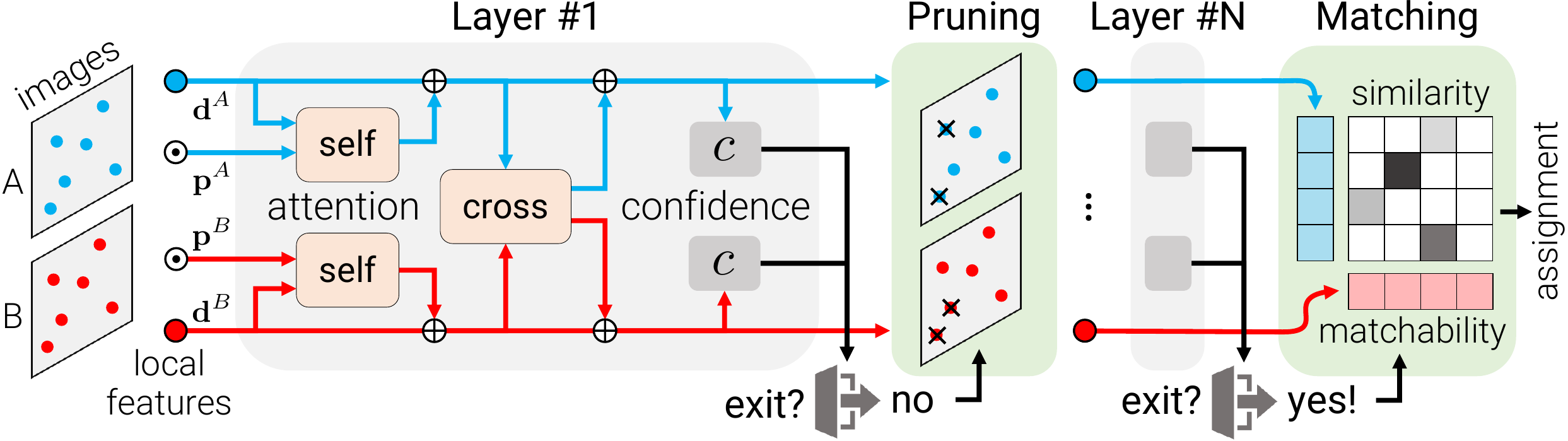}
    \caption{\textbf{The LightGlue architecture.}
    Given a pair of input local features ($\*d,\*p$), each layer augments the visual descriptors ($\color{red}\bullet$,$\color{blue}\bullet$) with context based on self- and cross-attention units with positional encoding $\odot$.
    A confidence classifier $c$ helps decide whether to stop the inference.
    If few points are confident, the inference proceeds to the next layer but we prune points that are confidently unmatchable.
    Once a confident state if reached, LightGlue predicts an assignment between points based on their pariwise similarity and unary matchability.
    }%
    \label{fig:architecture}%
\end{figure*}

\section{Related work}

\PAR{Matching images} that depict the same scene or object typically relies on local features, which are sparse keypoints each associated with a descriptor of its local appearance. 
While classical algorithms rely on hand-crafted criteria and gradient statistics~\cite{lowe2004distinctive, harris1988combined, bay2006surf, rosten2006machine}, much of the recent research has focused on designing Convolutional Neural Networks (CNNs) for both detection~\cite{yi2016lift,superpoint,dusmanu2019d2,revaud2019r2d2,tyszkiewicz2020disk} and description~\cite{HardNet,tian2019sosnet}.
Trained with challenging data, CNNs largely improve the accuracy and robustness of matching.
Local features now come in many flavors: some are better localized~\cite{lowe2004distinctive}, highly repeatable~\cite{superpoint}, cheap to store and match~\cite{Rublee2011ORB}, invariant to specific changes~\cite{Pautrat_2020_ECCV}, or ignore unreliable objects~\cite{tyszkiewicz2020disk}.\looseness=-1

Local features are then matched with a nearest neighbor search in descriptor space.
Because of non-matchable keypoints and imperfect descriptors, some correspondences are incorrect.
Those are filtered out by heuristics, like Lowe's ratio test~\cite{lowe2004distinctive} or the mutual check, inlier classifiers~\cite{moo2018learning,zhang2019learning}, and by robustly fitting geometric models~\cite{fischler1981random,cavalli2020handcrafted}.
This process requires extensive domain expertise and tuning and is prone to failure when conditions are too challenging.
These limitations are largely solved by deep matchers.

\PAR{Deep matchers} are deep networks trained to jointly match local features and reject outliers given an input image pair.
The first of its kind, SuperGlue~\cite{sarlin2020superglue} combines the expressive representations of Transformers~\cite{vaswani2017attention} with optimal transport~\cite{peyre2019computational} to solve a partial assignment problem.
It learns powerful priors about scene geometry and camera motion and is thus robust to extreme changes and generalizes well across data domains. 
Inheriting the limitations of early Transformers, SuperGlue is hard to train and its complexity grows quadratically with the number of keypoints.

Subsequent works make it more efficient by reducing the size of the attention mechanism.
They restrict it to a small set of seed matches~\cite{sgmnet} or within clusters of similar keypoints~\cite{clustergnn}.
This largely reduces the run time for large numbers of keypoints but yields no gains for smaller, standard input sizes.
This also impairs the robustness in the most challenging conditions, failing to reach the performance of the original SuperGlue model.
LightGlue instead brings large improvements for typical operating conditions, like in SLAM, without compromising on performance for any level of difficulty.
This is achieved by dynamically adapting the network size instead of reducing its overall capacity.

Conversely, dense matchers like LoFTR~\cite{sun2021loftr} and follow-ups~\cite{aspanformer,matchformer} match points distributed on dense grids rather than sparse locations.
This boosts the robustness to impressive levels but is generally much slower because it processes many more elements.
This limits the resolution of the input images and, in turn, the spatial accuracy of the correspondences.
While LightGlue operates on sparse inputs, we show that fair tuning and evaluation makes it competitive with dense matchers, for a fraction of the run time.

\PAR{Making Transformers efficient} has received significant attention following their success in language processing.
As the memory footprint of attention is a major limitation to handling long sequences, many works reduce it using linear formulations~\cite{linformer,linearatt,reformer} or bottleneck latent tokens~\cite{settransformer,perciever}.
This enables long-range context but can impair the performance for small input sizes.
Selective checkpointing~\cite{sqrtattention} reduces the memory footprint of attention and optimizing the memory access also drastically speeds it up~\cite{flashattention}.

Other, orthogonal works instead adaptively modulate the network depth by predicting whether the prediction of a token at a given layer is final or requires further computations~\cite{universaltransformer,DepthAdaptiveTransformer,calm} .
This is mostly inspired by adaptive schemes developed for CNNs by the vision community~\cite{branchynet,anytime_stereo,adaptive_dense,figurnov2017spatially,li2017not,verelst2020dynamic}.
In Transformers, the type of positional encoding has a large impact on the accuracy. 
While absolute sinusoidal~\cite{vaswani2017attention} or learned encodings~\cite{bert,gpt} were initially prevalent, recent works have studied relative encodings~\cite{shaw-etal-2018-self,roformer} to stabilize the training and better capture long-range dependencies.

LightGlue adapts some of these innovations to 2D feature matching and shows gains in both efficiency and accuracy.

\section{Fast feature matching}
\PAR{Problem formulation:}
LightGlue predicts a partial assignment between two sets of local features extracted from images $A$ and $B$, following SuperGlue.
Each local feature $i$ is composed of a 2D point position $\*p_i := (x, y)_i \in [0,1]^2$, normalized by the image size, and a visual descriptor $\*d_i \in \mathbb{R}^d$.
Images $A$ and $B$ have $M$ and $N$ local features, indexed by $\mathcal{A} := \{1, ..., M\}$ and $\mathcal{B} := \{1, ..., N\}$, respectively.

We design LightGlue to output a set of correspondences $\mathcal{M} = \{(i, j)\} \subset \mathcal{A}\times\mathcal{B}$.
Each point is matchable at least once, as it stems from a unique 3D point, and some keypoints are unmatchable, due to occlusion or non-repeatability.
As in previous works, we thus seek a soft partial assignment matrix $\*P\in[0,1]^{M\times N}$ between local features in $A$ and $B$, from which we can extract correspondences.

\PAR{Overview – Figure~\ref{fig:architecture}:}
LightGlue is made of a stack of $L$ identical layers that process the two sets jointly.
Each layer is composed of self- and cross-attention units that update the representation of each point.
A classifier then decides, at each layer, whether to halt the inference, thus avoiding unnecessary computations.
A lightweight head finally computes a partial assignment from the set of representations.

\subsection{Transformer backbone}
We associate each local feature $i$ in image $I \in \{A,B\}$ with a state $\*x^I_i\in\mathbb{R}^d$.
The state is initialized with the corresponding visual descriptor $\*x^I_i \leftarrow \*d^I_i$ and subsequently updated by each layer.
We define a layer as a succession of one self-attention and one cross-attention units. %

\PAR{Attention unit:}
In each unit, a Multi-Layer Perceptron (MLP) updates the state given a message $\*m^{I\leftarrow S}_i$ aggregated from a source image $S \in \{A,B\}$:
\begin{equation}
    \*x^I_i \leftarrow \*x^I_i + \operatorname{MLP}\left(\left[\*x^I_i\,|\,\*m^{I\leftarrow S}_i\right]\right) \enspace,
    \label{eq:update}
\end{equation}
where $[\cdot\,|\,\cdot]$ stacks two vectors.
This is computed for all points in both images in parallel.
In a self-attention unit, each image $I$ pulls information from points of the same image and thus $S=I$.
In a cross-attention unit, each image pulls information from the other image and $S = \{A,B\}\backslash I$.

The message is computed by an attention mechanism as the weighted average of all states $j$ of image $S$:
\begin{equation}
    \*m^{I\leftarrow S}_i = \sum_{j\in\mathcal{S}} \underset{k\in\mathcal{S}}{\operatorname{Softmax}}\left(a_{ik}^{IS}\right)_j
    \*W \*x_j^S \enspace,
\label{eq:attention}
\end{equation}
where $\*W$ is a projection matrix
and $a_{ij}^{IS}$ is an attention score between points $i$ and $j$ of images $I$ and $S$.
How this score is computed differs for self- and cross-attention units.

\PAR{Self-attention:}
Each point attends to all points of the same image.
We perform the same following steps for each image~$I$ and thus drop the superscript $I$ for clarity.
For each point $i$, the current state $\*x_i$ is first decomposed into key and query vectors $\*k_i$ and $\*q_i$ via \emph{different} linear transformations.
We then define the attention score between points $i$ and $j$ as
\begin{equation}
    a_{ij} = \*q_i^\top\,\*R\!\left(\*p_j - \*p_i\right)\,\*k_j\enspace,
\end{equation}
where $\*R\!\left(\cdot\right)\in\mathbb{R}^{d\times d}$ is a rotary encoding~\cite{roformer} of the relative position between the points.
We partition the space into $d/2$ 2D subspaces and rotate each of them by an angle corresponding, following Fourier Features~\cite{fourierfeatures}, to the projection onto a learned basis~ $\*b_k\in\mathbb{R}^2$:
\begin{equation}
    \*R\!\left(\*p\right) = 
    \left(
    \begin{smallmatrix}
    \hat{\*R}(\*b_1^\top\*p) && \*0\\
    &\ddots\\
    \*0 && \hat{\*R}(\*b_{d\!/\!2}^\top\*p)
    \end{smallmatrix}
    \right)
    \!, 
    \hat{\*R}(\theta) = \left(
    \begin{smallmatrix}
    \cos\theta&-\sin\theta\\
    \sin\theta&\cos\theta
    \end{smallmatrix}
    \right).
\end{equation}

Positional encoding is a critical part of attention as it allows addressing different elements based on their position.
We note that, in projective camera geometry, the position of visual observations is equivariant w.r.t.\ a translation of the camera within the image plane:
2D points that stem from 3D points on the same fronto-parallel plane are translated in an identical way and their relative distance remains constant.
This calls for an encoding that only captures the relative but not the absolute position of points.

The rotary encoding~\cite{roformer} enables the model to retrieve points $j$ that are located at a learned relative position from $i$.
The positional encoding is not applied to the value $\*v_j$ and thus does not spill into the state $\*x_i$.
The encoding is identical for all layers and is thus computed once and cached.

\PAR{Cross-attention:}
Each point in $I$ attends to all points of the other image $S$.
We compute a key $\*k_i$ for each element but no query.
This allows to express the score as
\begin{equation}
    a_{ij}^{IS} = \*k_i^I{}^\top\*k_j^S \overset{!}{=} a_{ji}^{SI}\enspace.
\end{equation}
We thus need to compute the similarity only once for both $I\!\leftarrow\!S$ and $S\!\leftarrow\!I$ messages.
This trick has been previously referred to as \emph{bidirectional} attention~\cite{bidirectional}.
Since this step is expensive, with a complexity of $O(N\!M\!d)$, it saves a significant factor of 2.
We do not add any positional information as relative positions are not meaningful across images.

\subsection{Correspondence prediction}
We design a lightweight head that predicts an assignment given the updated state at any layer.

\PAR{Assignment scores:}
We first compute a pairwise score matrix $\*S\in\mathbb{R}^{M\times N}$ between the points of both images:
\begin{equation}
\*S_{ij} = \operatorname{Linear}\left(\*x_i^A\right){}^\top\operatorname{Linear}\left(\*x_j^B\right)
\quad\forall (i, j) \in \mathcal{A} \times \mathcal{B},
\label{eq:assignment}
\end{equation}
where $\operatorname{Linear}(\cdot)$ is a learned linear transformation with bias.
This score encodes the affinity of each pair of points to be in correspondence, \ie 2D projections of the same 3D point.
We also compute, for each point, a matchability score as
\begin{equation}
\sigma_i = \operatorname{Sigmoid}\left(\operatorname{Linear}(\*x_i)\right)\in[0,1]\enspace.
\end{equation}
This score encodes the likelihood of $i$ to have a corresponding point.
A point that is not detected in the other image, \eg~when occluded, is not matchable and thus has $\sigma_i\rightarrow0$.

\PAR{Correspondences:}
We combine both similarity and matchability scores into a soft partial assignment matrix $\*P$ as
\begin{equation}
    \*P_{ij} =
    \sigma_i^A\;
    \sigma_j^B\;
    \underset{k\in\mathcal{A}}{\operatorname{Softmax}}(\*S_{kj})_i\;
    \underset{k\in\mathcal{B}}{\operatorname{Softmax}}(\*S_{ik})_j\enspace.
\end{equation}
A pair of points $(i,j)$ yields a correspondence when both points are predicted as matchable and when their similarity is higher than any other point in both images.
We select pairs for which $\*P_{ij}$ is larger than a threshold $\tau$ and than any other element along both its row and column.

\subsection{Adaptive depth and width}
We add two mechanisms that avoid unnecessary computations and save inference time: 
i)~we reduce the number of layers depending on the difficulty of the input image pair;
ii)~we prune out points that are confidently rejected early.

\begin{figure}[t]
    \centering
    \includegraphics[width=\linewidth]{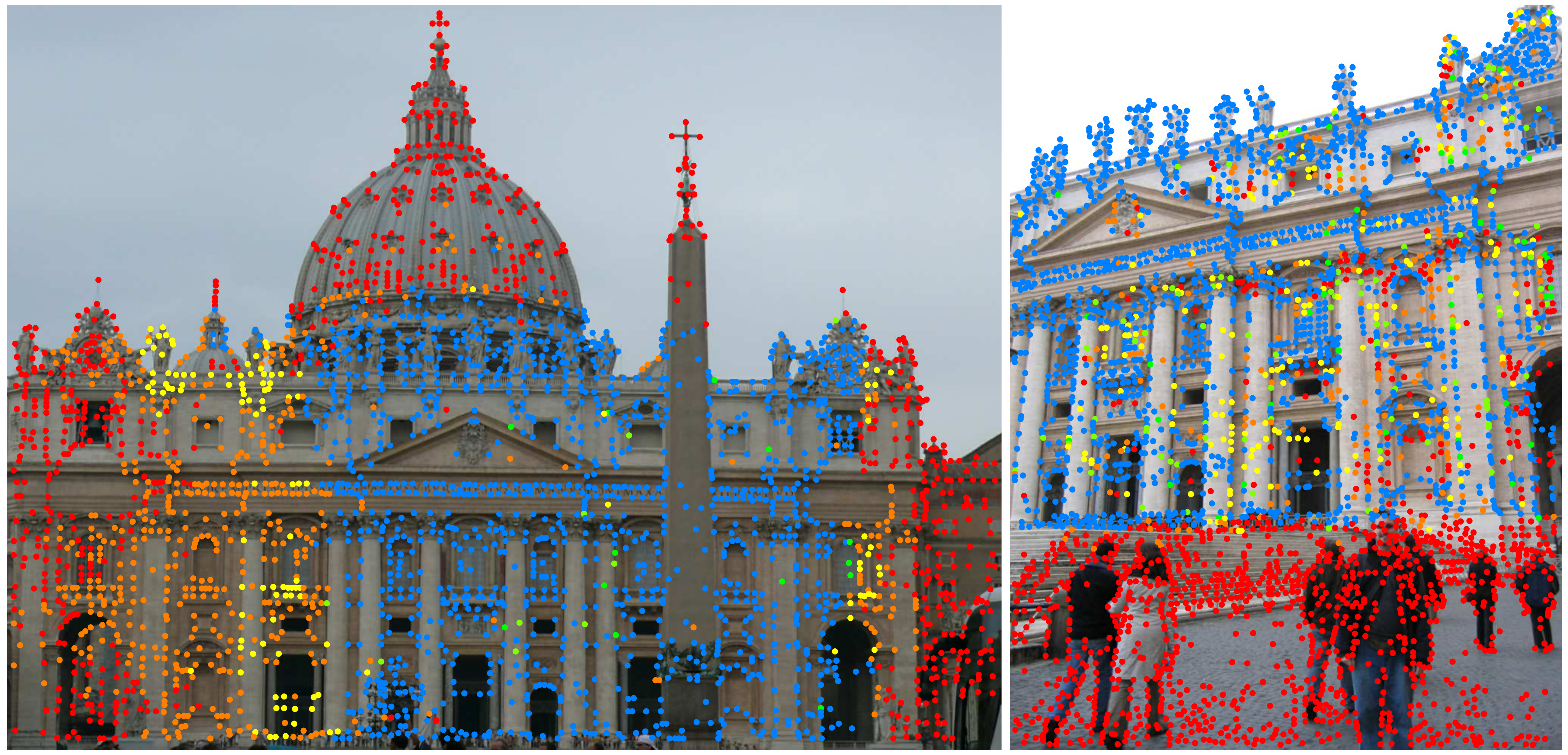}
    \caption{\textbf{Point pruning.} 
    As LigthGlue aggregates context, 
    it can find out early that some points ($\color{red}\bullet$) are unmatchable and thus exclude them from subsequent layers.
    Other, non-repeatable points are excluded in later layers: ${\color{orange}\bullet} \rightarrow {\color{yellow}\bullet}\rightarrow {\color{green}\bullet}$.
    This reduces the inference time and the search space ($\color{blue}\bullet$) to ultimately find good matches fast.
    }%
    \label{fig:confidence-example}%
\end{figure}

\PAR{Confidence classifier:}
The backbone of LightGlue augments input visual descriptors with context.
These are often reliable if the image pair is easy, \ie~has high visual overlap and little appearance changes.
In such case, predictions from early layers are confident and identical to those of late layers.
We can then output these predictions and halt the inference.

At the end of each layer, LightGlue infers the confidence of the predicted assignment of each point:
\begin{equation}
    c_i = \operatorname{Sigmoid}\left(\operatorname{MLP}(\*x_i)\right)\in[0,1]\enspace.
    \label{eq:exit-classifier}
\end{equation}
A higher value indicates that the representation of $i$ is reliable and final -- it is confidently either matched or unmatchable.
This is inspired by multiple works that successfully apply this strategy to language and vision tasks~\cite{calm,DepthAdaptiveTransformer,branchynet,anytime_stereo,adaptive_dense}.
The compact MLP adds only 2\% of inference time in the worst case but most often saves much more.

\PAR{Exit criterion:}
For a given layer $\ell$, a point is deemed confident if $c_i > \lambda_\ell$.
We halt the inference if a sufficient ratio $\alpha$ of all points is confident:
\begin{equation}
    \mathrm{exit} = \left(\frac{1}{N\!+\!M}\sum_{I\in\{A,B\}}\sum_{i\in\mathcal{I}}\ \llbracket c^I_i > \lambda_\ell\rrbracket\right) > \alpha
    \enspace.
\end{equation}
We observe, as in \cite{calm}, that the classifier itself is less confident in early layers.
We thus decay $\lambda_\ell$ throughout the layers based on the validation accuracy of each classifier.
The exit threshold $\alpha$ directly controls the trade-off between accuracy and inference time.

\PAR{Point pruning:}
When the exit criterion is not met, points that are predicted as both confident and unmatchable are unlikely to aid the matching of other points in subsequent layers.
Such points are for example in areas that are clearly not covisible across the images.
We therefore discard them at each layer and feed only the remaining points to the next one.
This significantly reduces computation, given the quadratic complexity of attention, and does not impact the accuracy.

\subsection{Supervision}
We train LightGlue in two stages: we first train it to predict correspondences and only after train the confidence classifier.
The latter thus does not impact the accuracy at the final layer or the convergence of the training.

\PAR{Correspondences:}
We supervise the assignment matrix $\*P$ with ground truth labels estimated from two-view transformations.
Given a homography or pixel-wise depth and a relative pose, we wrap points from $A$ to $B$ and conversely.
Ground truth matches $\mathcal{M}$ are pairs of points with a low reprojection error in both images and a consistent depth.
Some points $\bar{\mathcal{A}}\subseteq\mathcal{A}$ and $\bar{\mathcal{B}}\subseteq\mathcal{B}$ are labeled as unmatchable when their reprojection or depth errors are sufficiently large with all other points.
We then minimize the log-likelihood of the assignment predicted at each layer $\ell$, 
pushing LightGlue to predict correct correspondences early:
\begin{equation}
    \begin{split}
    \text{loss} =
     -\frac{1}{L}\sum_{\ell}\Bigg(
     &\frac{1}{|\mathcal{M}|}\sum_{(i, j) \in \mathcal M} \log {}^\ell\*P_{ij}\\
     &+ \frac{1}{2|\bar{\mathcal{A}}|}\sum_{i \in \bar{\mathcal{A}}} \log \left(1 - {}^\ell\sigma_i^A\right)\\
     &+ \frac{1}{2|\bar{\mathcal{B}}|}\sum_{j \in \bar{\mathcal{B}}} \log \left(1 - {}^\ell\sigma_j^B\right)\Bigg)\enspace.
\end{split}
\end{equation}
The loss is balanced between positive and negative labels.

\PAR{Confidence classifier:}
We then train the MLP of Eq.~(\ref{eq:exit-classifier}) to predict whether the prediction of each layer is identical to the final one.
Let ${}^{\ell}m^A_i \in \mathcal{B}\cup\{\bullet\}$ be the index of the point in $B$ matched to $i$ at layer $\ell$, with ${}^{\ell}m^A_i{=}\,\bullet\,$ if $i$ is unmatchable.
The ground truth binary label of each point is $\llbracket{}^{\ell}m^A_i = {}^{L}m^A_i\rrbracket$ and identically for $B$.
We then minimize the binary cross-entropy of the classifiers of layers $\ell\in\{1, ..., L\!-\!1\}$.

\subsection{Comparison with SuperGlue}
LightGlue is inspired by SuperGlue but differs in aspects critical to its accuracy, efficiency, and ease of training.

\PAR{Positional encoding:}
SuperGlue encodes the absolute point positions with an MLP and fuses them early with the descriptors.
We observed that the model tends to forget this positional information throughout the layers.
LightGlue instead relies on a relative encoding that is better comparable across images and is added in each self-attention unit.
This makes it easier to leverage the positions and improves the accuracy of deeper layers.

\PAR{Prediction head:}
SuperGlue predicts an assignment by solving a differentiable optimal transport problem using the Sinkhorn algorithm~\cite{sinkhorn1967concerning, peyre2019computational}.
It consists in many iterations of row-wise and column-wise normalization, which is expensive in terms of both compute and memory.
SuperGlue adds a dustbin to reject unmatchable points.
We found that the dustbin entangles the similarity score of all points and thus yields suboptimal training dynamics.
LightGlue disentangles similarity and matchability, which are much more efficient to predict. 
This also yields cleaner gradients.

\PAR{Deep supervision:}
Because of how expensive Sinkhorn is, SuperGlue cannot make predictions after each layer and is supervised only at the last one.
The lighter head of LightGlue makes it possible to predict an assignment at each layer and to supervise it.
This speeds up the convergence and enables exiting the inference after any layer, which is key to the efficiency gains of LightGlue.

\begin{figure}[t]
    \centering
    \includegraphics[width=\linewidth]{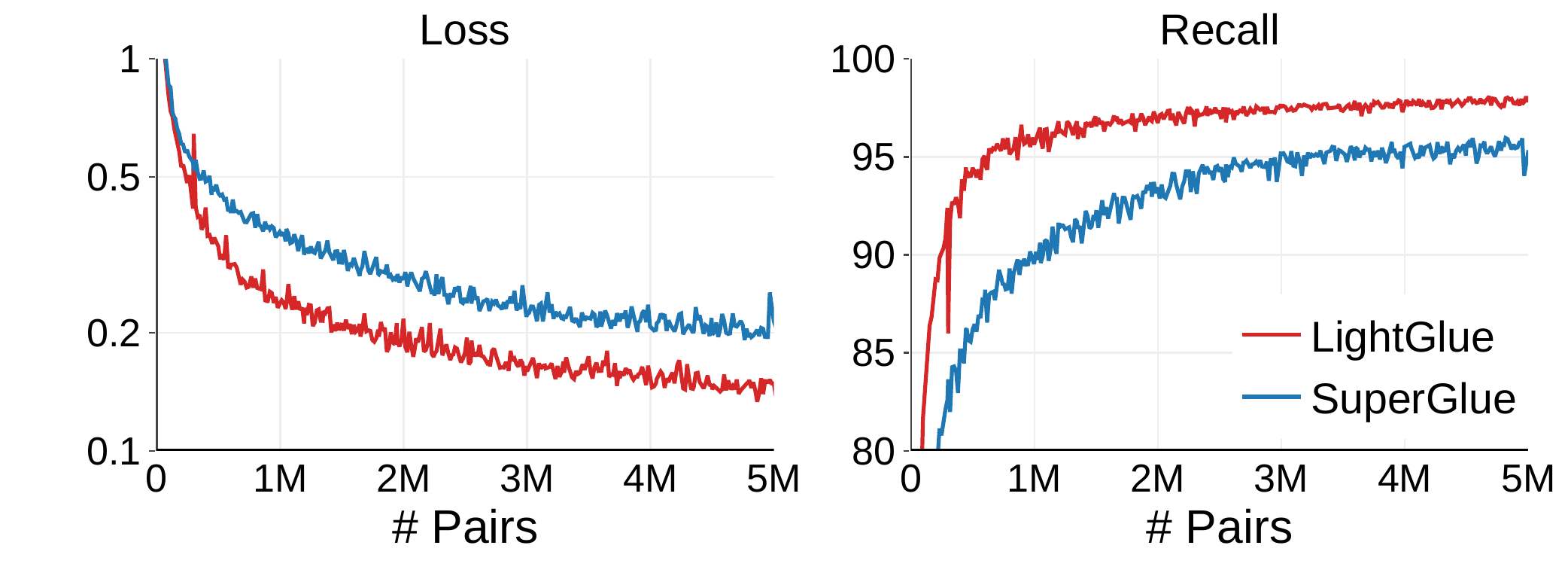}
    \caption{\textbf{Ease of training.}
    The LightGlue architecture vastly improves the speed of convergence of the pre-training on synthetic homographies.
    After 5M image pairs (only 2 GPU-days), LighGlue achieves -33\% loss at the final layer and +4\% match recall. 
    SuperGlue requires over 7 days of training to reach a similar accuracy.
    }%
    \label{fig:training}%
\end{figure}

\section{Details that matter}

\PAR{Recipe:}
LightGlue follows the supervised training setup of SuperGlue.
We first pre-train the model with synthetic homographies sampled from 1M images~\cite{radenovic2018revisiting}.
Such augmentations provide full and noise-free supervision but require careful tuning.
LightGlue is then fine-tuned with the MegaDepth dataset~\cite{li2018megadepth}, which includes 1M crowd-sourced images depicting 196 tourism landmarks, with camera calibration and poses recovered by SfM and dense depth by multi-view stereo.
Because large models easily overfit to such distinctive scenes, the pre-training is critical to the generalization of the model but was omitted in recent follow-ups~\cite{sgmnet,clustergnn}.

\PAR{Training tricks:}
While the LightGlue architecture improves the training speed, stability, and accuracy, we found that some details have a large impact too.
Figure~\ref{fig:training} shows that this reduces the resources required to train a model compared to SuperGlue.
This lowers the cost of training and makes deep matchers more accessible to the broader community.

Since the depth maps of MegaDepth are often incomplete, we also label points with a large epipolar error as unmatchable.
Carefully tuning and annealing the learning rate boosts the accuracy.
Training with more points also does: we use 2k per image instead of 1k.
The batch size matters: we use gradient checkpointing~\cite{checkpointing} and mixed-precision to fit 32 image pairs on a single GPU with 24GB VRAM.

\PAR{Implementation details:}
LighGlue has $L{=}9$ layers.
Each attention unit has 4 heads.
All representations have dimension $d{=}256$.
Throughout the paper, run-time numbers labeled as \emph{optimized} use an efficient implementation of self-attention~\cite{flashattention}.
More details are given in the \supp.

We train LightGlue with both SuperPoint~\cite{superpoint} and SIFT~\cite{lowe2004distinctive} local features but it is compatible with any other type.
When fine-tuning the model on MegaDepth~\cite{li2018megadepth}, we use the data splits of Sun~\etal~\cite{sun2021loftr} to avoid training on scenes included in the Image Matching Challenge~\cite{Jin2020}.

\section{Experiments}
We evaluate LightGlue for the tasks of homography estimation, relative pose estimation, and visual localization.
We also analyze the impacts of our design decisions.

\subsection{Homography estimation}
We evaluate the quality of correspondences estimated by LightGlue on planar scenes of the HPatches~\cite{balntas2017hpatches} dataset.
This dataset is composed of sequences of 5 image pairs, each under either illumination or viewpoint changes.

\PAR{Setup:} 
Following SuperGlue~\cite{sarlin2020superglue}, we report the precision and recall compared to GT matches at a reprojection error of 3px.
We also evaluate the accuracy of homographies estimated from the correspondences using robust and non-robust solvers: RANSAC~\cite{fischler1981random} and the weighted DLT~\cite{hartley2003multiple}.
For each image pair, we compute the mean reprojection error of the four image corners and report the area under the cumulative error curve (AUC) up to values of 1px and 5px.
Following best practices in benchmarking~\cite{Jin2020} and unlike past works~\cite{sarlin2020superglue, sun2021loftr},
we use a state-of-the-art robust estimator~\cite{barath2019magsac}
and extensively tune the inlier threshold for each method separately.
We then report the highest scoring results.

\begin{table}[t]
\centering
\footnotesize{\setlength\tabcolsep{3.0pt}%
\begin{tabular}{clccccccc}
    \toprule
    \multicolumn{2}{c}{\multirow{2}{*}[-.4em]{features + matcher}}&
    \multirow{2}{*}[-.4em]{R} &
    \multirow{2}{*}[-.4em]{P} &
    \multicolumn{2}{c}{AUC - RANSAC} & 
    \multicolumn{2}{c}{AUC - DLT}\\
    \cmidrule(lr){5-6}
    \cmidrule(lr){7-8}
    &&&& @1px & @5px & @1px & @5px  \\
    \midrule
    dense & LoFTR & - & 92.7 & 41.5 & 78.8 & 38.5 & 70.6 \\
    \midrule[0.2pt]
    \multirow{3}{*}{\begin{sideways}SuperPoint\end{sideways}}
    & NN+mutual & 72.7 & 67.2 & 35.0 & 75.3 & \00.0 & \02.0 \\
    & SuperGlue & 94.9 & 87.4 & 38.3 & 79.3 & 33.8 & 76.7 \\
    & SGMNet & \b{95.5} & 83.0 & \b{38.6} & 79.0 & 31.7 & 76.0 \\
    & \b{LightGlue} & 94.3 & \b{88.9} & 38.3 & \b{79.6} & \b{35.9} & \b{78.6} \\
    \bottomrule
\end{tabular}}
\caption{\textbf{Homography estimation on HPatches.}
LightGlue yields better correspondences than sparse matchers, with the highest precision (P) and a high recall (R).
This results in accurate homographies when estimated by RANSAC or even a faster least-squares solver (DLT).
LightGlue is competitive with dense matchers like LoFTR.
}%
\label{tab:hpatches}
\end{table}

\PAR{Baselines:}
We follow the setup of~\cite{sun2021loftr} and resize all images such that their smaller dimension is equal to 480 pixels.
We evaluate sparse matchers with 1024 local features extracted by SuperPoint~\cite{superpoint}.
We compare LightGlue against nearest-neighbor matching with mutual check and the deep matchers SuperGlue~\cite{sarlin2020superglue} and SGMNet~\cite{sgmnet}.
We use the official models trained on outdoor datasets~\cite{li2018megadepth,shen2018mirror}.
For reference, we also evaluate the dense matcher LoFTR~\cite{sun2021loftr}, selecting only the top 1024 predicted matches for the sake of fairness.

\PAR{Results:}
Table~\ref{tab:hpatches} shows that 
LightGlue yields correspondences with higher precision than and similar recall to SuperGlue and SGMNet.
When estimating homographies with DLT, this results in much more accurate estimates than with other matchers.
LightGlue thus makes DLT, a simple solver, competitive with the expensive and slower MAGSAC~\cite{barath2019magsac}.
At a coarse threshold of 5px, LightGlue is also more accurate than LoFTR despite being constrained by sparse keypoints.

\subsection{Relative pose estimation}
\label{sec:rel-pose-outdoor}
We evaluate LightGlue for pose estimation in outdoor scenes that exhibit strong occlusion and challenging lighting and structural changes.

\PAR{Setup:}
We use image pairs from the MegaDepth-1500 test set following the evaluation of ~\cite{sun2021loftr}.
The test set contains 1500 image pairs from two popular phototourism destinations: St. Peters Square and Reichstag. The data was collected in a way that the difficulty is balanced based on visual overlap. We evaluate our method on the downstream task of relative pose estimation.

We estimate an essential matrix both with vanilla RANSAC and LO-RANSAC~\cite{PoseLib}, respectively, and decompose them into a rotation and a translation.
The inlier threshold is tuned for each approach on the test data -- we think that this makes the comparison more fair as we do not evaluate RANSAC itself.
We compute the pose error as the maximum angular error in rotation and translation and we report its AUC at 5°, 10°, and 20°.

\begin{table}[t]
\centering
\footnotesize{\setlength\tabcolsep{4.7pt}%
\begin{tabular}{rlccc}
    \toprule
    \multicolumn{2}{c}{\multirow{2}{1.5cm}[-.4em]{features + matcher}}
    & RANSAC AUC & LO-RANSAC AUC
    &\multirow{2}{0.5cm}[-.4em]{time (ms)}
    \\
     \cmidrule(lr){3-4}
    & & \multicolumn{2}{c}{5° / 10° / 20°} & \\
    \midrule
    \multirow{3}{*}{\begin{sideways}dense\end{sideways}}
    
    & LoFTR & 52.8 / 69.2 / 81.2 & 66.4 / 78.6 / 86.5 & 181 \\
    & MatchFormer & 53.3 / 69.7 / 81.8 & 66.5 / 78.9 / 87.5 & 388 \\
    & ASpanFormer & \b{55.3} / \b{71.5} / \b{83.1} & \b{69.4} / \b{81.1} / \b{88.9} & 369 \\ %
    
    \midrule[0.2pt]
    \multirow{2}{*}{\begin{sideways}DISK\end{sideways}}
    & NN+ratio & 38.1 / 55.4 / 69.6 & 57.2 / 69.5 / 78.6 & 7.4 \\
    & \b{LightGlue} & \b{43.5} / \b{61.0} / \b{75.3} & \b{61.3} / \b{74.3} / \b{83.8} & 44.5 \\
    \midrule[0.2pt]
    \multirow{5}{*}{\begin{sideways}SuperPoint\end{sideways}}
    & NN+mutual & 31.7 / 46.8 / 60.1 & 51.0 / 54.1 / 73.6 & 5.7 \\
    & SuperGlue & 49.7 / 67.1 / \b{80.6} & 65.8 / 78.7 / 87.5 & 70.0 \\
    & SGMNet & 43.2 / 61.6 / 75.6  & 59.8 / 74.1 / 83.9 & 73.8 \\
    & \b{LightGlue} & \b{49.9} / 67.0 / 80.1 & \b{66.7} / \b{79.3} / \b{87.9} & 44.2 \\
    & \b{\cornerarrow adaptive} & 49.4 / \b{67.2} / 80.1 & 66.3 / 79.0 / \b{87.9} & 31.4 \\
    
    \bottomrule
\end{tabular}}
\caption{\textbf{Relative pose estimation.}
On the MegaDepth1500 dataset, LightGlue predicts more precise correspondences with higher pose accuracy (AUC), and speed than existing sparse matchers.
It is competitive with dense matchers for a fraction of the inference time, and even outperforms LoFTR and MatchFormer with the superior LO-RANSAC estimator.
The adaptive scheme greatly reduces the run time for only a minor loss of accuracy.
}%
\label{tab:megadepth}
\end{table}

\PAR{Baselines:}
We extract 2048 local features per images, each resized such that its larger dimension is 1600 pixels.
With SuperPoint~\cite{superpoint} features, we compare LightGlue to nearest-neighbor matching with mutual check and to the official implementations of SuperGlue~\cite{sarlin2020superglue} and SGMNet~\cite{sgmnet}.
With DISK~\cite{tyszkiewicz2020disk} we only evaluate against its own strong baseline, as no other trained matcher with DISK is publicly available.

We also evaluate the recent, dense deep matchers LoFTR~\cite{sun2021loftr}, MatchFormer~\cite{matchformer}, and ASpanFormer~\cite{aspanformer}.
We carefully follow their respective evaluation setups and resize the input images such that their largest dimension is 840 pixels (LoFTR, MatchFormer) or 1152 pixels (ASpanFormer).
Larger images would improve their accuracy, as with sparse features, but would incur prohibitive and unpractical run time and memory requirements.

\PAR{Results:}
Table~\ref{tab:megadepth} shows that LightGlue largely outperforms the existing approaches SuperGlue and SGMNet on SuperPoint features, and can greatly improve the matching accuracy over DISK local features.
It yields better correspondences and more accurate relative poses and reduces the inference time by 30\%.
LightGlue typically predicts slightly fewer matches than SuperGlue but those are more accurate.
By detecting confident predictions early in the model, 
the adaptive variant is over 2$\times$ faster than SuperGlue and SGMNet and still more accurate.
With a carefully tuned LO-RANSAC~\cite{PoseLib}, LightGlue can achieve higher accuracy than some popular dense matcher which are between 5 and 11 times slower.
Among the evaluated dense matchers, ASPANFormer is the most accurate.
Considering trade-off between accuracy and speed, LightGlue outperforms all approaches by a large margin.

\subsection{Outdoor visual localization}
\PAR{Setup:} 
We evaluate long-term visual localization in challenging conditions using the large-scale Aachen Day-Night benchmark~\cite{sattler2018benchmarking}.
We follow the Hierarchical Localization framework with the hloc toolbox~\cite{sarlin2019coarse}.
We first triangulate a sparse 3D point cloud from the 4328 daytime reference images, with known poses and calibration, using COLMAP~\cite{schoenberger2016sfm}.
For each of the 824 daytime and 98 nighttime queries, we retrieve 50 images with NetVLAD~\cite{arandjelovic2016netvlad}, match each of them, and estimate a camera pose with RANSAC and a Perspective-n-Point solver.
We report the pose recall at multiple thresholds
and the average throughput of the matching step during both mapping and localization.

\begin{table}[t]
\centering
\footnotesize{\setlength\tabcolsep{4.7pt}%
\begin{tabular}{lccccc}
    \toprule
    \multirow{2}{1.3cm}[-.4em]{SuperPoint + matcher}
    & Day & & Night
    & \multirow{2}{1.0cm}[-.4em]{\centering pairs per second}\\
    \cmidrule(lr){2-4}
    & \multicolumn{3}{c}{(0.25m,2°) / (0.5m,5°) / (1.0m,10°)}\\
    \midrule
    SuperGlue & 88.2 / \b{95.5} / \b{98.7} & & 86.7 / 92.9 / \b{100} & 6.5 \\
    SGMNet & 86.8 / 94.2 / 97.7 & & 83.7 / 91.8 / 99.0 & 10.2 \\
    ClusterGNN & \b{89.4} / \b{95.5} / 98.5 & & 81.6 / \b{93.9} / \b{100} & 13* \\
    \b{LightGlue} & 89.2 / 95.4 / 98.5 & & \b{87.8} / \b{93.9} / \b{100} & \b{17.2 / \emph{26.1}} \\
    \bottomrule
\end{tabular}}
\caption{\textbf{Outdoor visual localization.} On the Aachen Day-Night dataset,
LightGlue performs on par with SuperGlue but runs 2.5$\times$ faster, 4$\times$ when \emph{optimized}.
SGMNet and ClusterGNN are both slower and less robust on night-time images (*approximation).
}%
\label{tab:aachen}
\end{table}

\PAR{Baselines:} 
We extract up to 4096 features with SuperPoint and match them with SuperGlue, SGMNet~\cite{sgmnet}, ClusterGNN~\cite{clustergnn}, and LightGlue with adaptive depth and width.
Since the implementation of ClusterGNN is not publicly available, we report the accuracy found in the original paper and the time estimates kindly provided by the authors.

\PAR{Results:}
Table~\ref{tab:aachen} shows that LightGlue reaches a similar accuracy as SuperGlue but at a 2.5$\times$ higher throughput.
The \emph{optimized} variant, which leverages an efficient self-attention~\cite{flashattention}, increases the throughput by 4$\times$.
LightGlue thus matches up to 4096 keypoints in real time.

\subsection{Insights}

\PAR{Ablation study:}
We validate our design decisions by evaluating LightGlue after its pre-training on the challenging synthetic homography dataset with extreme photometric augmentations.
We train different variants with SuperPoint features and 5M samples, all within 4 GPU-days.
We create a test set from the same augmentations applied to images unseen during training.
We extract 512 keypoints from each.
We also compare against SuperGlue, which we train with the same setup. More details are provided in the \supp.

We report the ablation results in Table~\ref{tab:ablation}. Compared to SuperGlue, LightGlue converges significantly faster, and achieves +4\% recall and +12\% precision. Note that SuperGlue can achieve similar accuracies as LightGlue with a long-enough training, but the improved convergence makes it much more practical to train on new data.

\begin{figure}[t]
    \centering
    \includegraphics[width=0.95\linewidth]{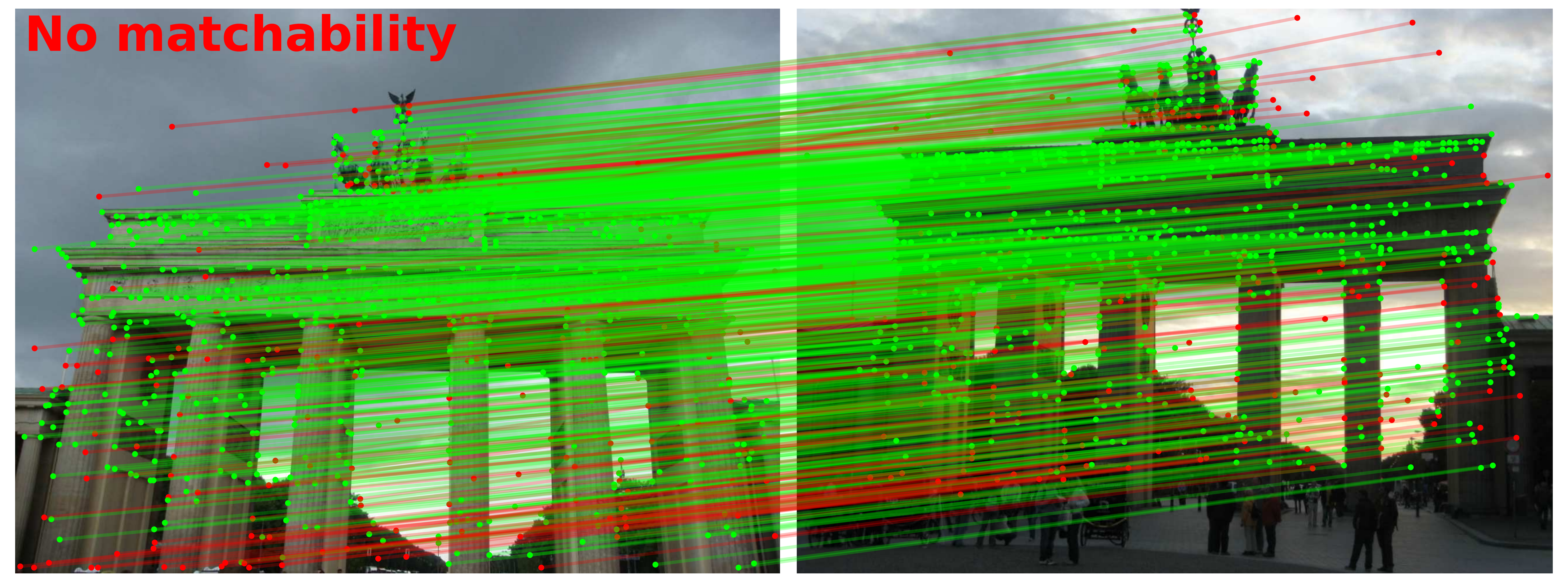}

    \includegraphics[width=0.95\linewidth]{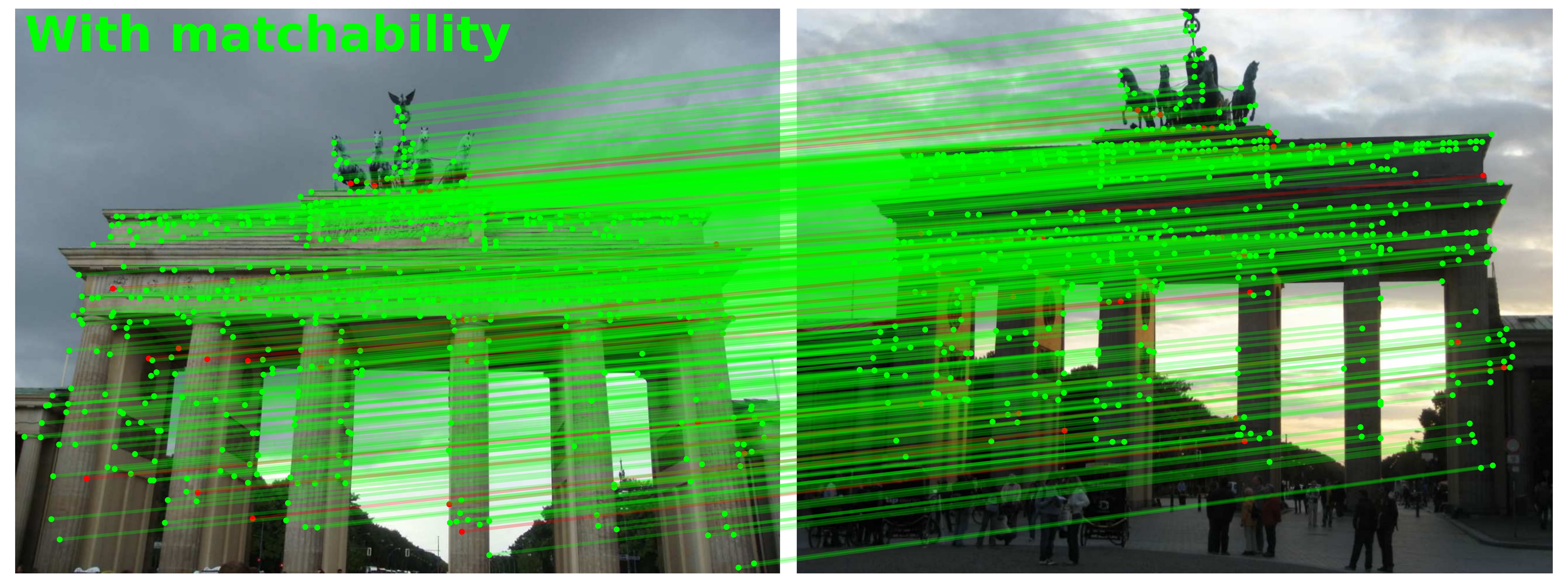}
    \caption{\textbf{Benefit of the matchability.}
    The matchability helps filter out outliers (\red{red}) that are visually similar, retaining only inlier correspondences (\green{green}).}%
    \label{fig:matchability}%
\end{figure}

Without the matchability classifier, the network loses its ability to discriminate between good and bad matches, as shown in Figure~\ref{fig:matchability}.
Intuitively, the similarity matrix proposes many likely matches while the matchability filters incorrect proposals.
Thus, our partial assignment can be viewed as an elegant fusion of mutual nearest neighbor search and a learned inlier classifier~\cite{moo2018learning, zhang2019learning}.
This is significantly faster than solving the optimal transport problem of SuperGlue.

Replacing learned absolute positional encoding with rotary embeddings improves the accuracy, with a minor penalty on run time from rotating queries and keys at each self-attention layer. Using relative positions, LightGlue learns to match geometric patterns across images.
Reminding the network about positions at each layer improves the robustness of the network, resulting in +2\% precision.

Bidirectional cross-attention is equally accurate as standard cross-attention, but saves 20\% run time by only computing the similarity matrix once. Currently, the bottleneck is computing the softmax along two dimensions. With a dedicated bidirectional softmax kernel, plenty of redundant computations could be avoided.

Using deep supervision, also intermediate layers have meaningful outputs. Already after 5 layers, the network can predict robust matches, achieving $>90\%$ recall. In the final layers, the network focuses on rejecting outliers, thus improving the match precision.

\begin{table}[t]
\centering
\footnotesize{\setlength\tabcolsep{6.0pt}%
\begin{tabular}{lccc}
    \toprule
    architecture & precision & recall & time (ms) \\
    \midrule
    SuperGlue & 74.6 & 90.5 & 29.1 \\
    \midrule[0.2pt]
    \b{LightGlue} (full) & \b{86.8} & 96.3 & 19.4 \\
    \cornerarrow a) no matchability & 67.4 & \b{97.0} & 18.9 \\
    \cornerarrow b) absolute positions & 84.2 & 94.7 & 18.7 \\
    \cornerarrow c) full cross-attention & 86.6 & 96.1 & 22.8 \\
    \cornerarrow d) early layer (\#5/9) & 78.1 & 92.7 & \b{11.9} \\
    \bottomrule
\end{tabular}}
\caption{\textbf{Ablation study on synthetic homographies.}
a-b) Both matchability and positional encoding improve the accuracy without impact on the time.
c) The bidirectional cross-attention is faster without drop of accuracy.
d) Thanks to the deep supervision, early layers yield good predictions on pairs with low difficulty.
}%
\label{tab:ablation}
\end{table}

\begin{table}[t]
\centering
\footnotesize{\setlength\tabcolsep{4.0pt}%
\begin{tabular}{lccccccc}
    \toprule
    \multirow{2}{1.3cm}[-.4em]{metric}
    & \multicolumn{3}{c}{difficulty}
    & \multirow{2}{1.0cm}[-.4em]{\centering average}\\
    \cmidrule(lr){2-4}
    & easy & medium & hard\\
    \midrule
    average index of stopping layer $\downarrow$ & 4.7 & 5.5 & 6.9 & 5.7 \\
    ratio of unmatchable points (\%) 	$\uparrow$ & 19.8 & 23.4 & 27.9 & 23.7 \\
    speedup over non-adaptive $\uparrow$ & 1.86 & 1.33 & 1.16 & 1.45 \\
    \bottomrule
\end{tabular}}
\caption{\textbf{Impact of adaptive depth and width.} Early stopping helps most on smaller scenes, where the network stops after just half the layers. On harder scenes, the network requires more layers to converge, but smaller view overlap between image pairs allows the network to more aggressively prune the width of the network. Overall, adaptive depth- and width- pruning reduces the run time by 33\% and is particularly effective on easy pairs.
}%
\label{tab:adaptive}
\end{table}

\begin{figure}[t]
    \centering
    \includegraphics[width=\linewidth]{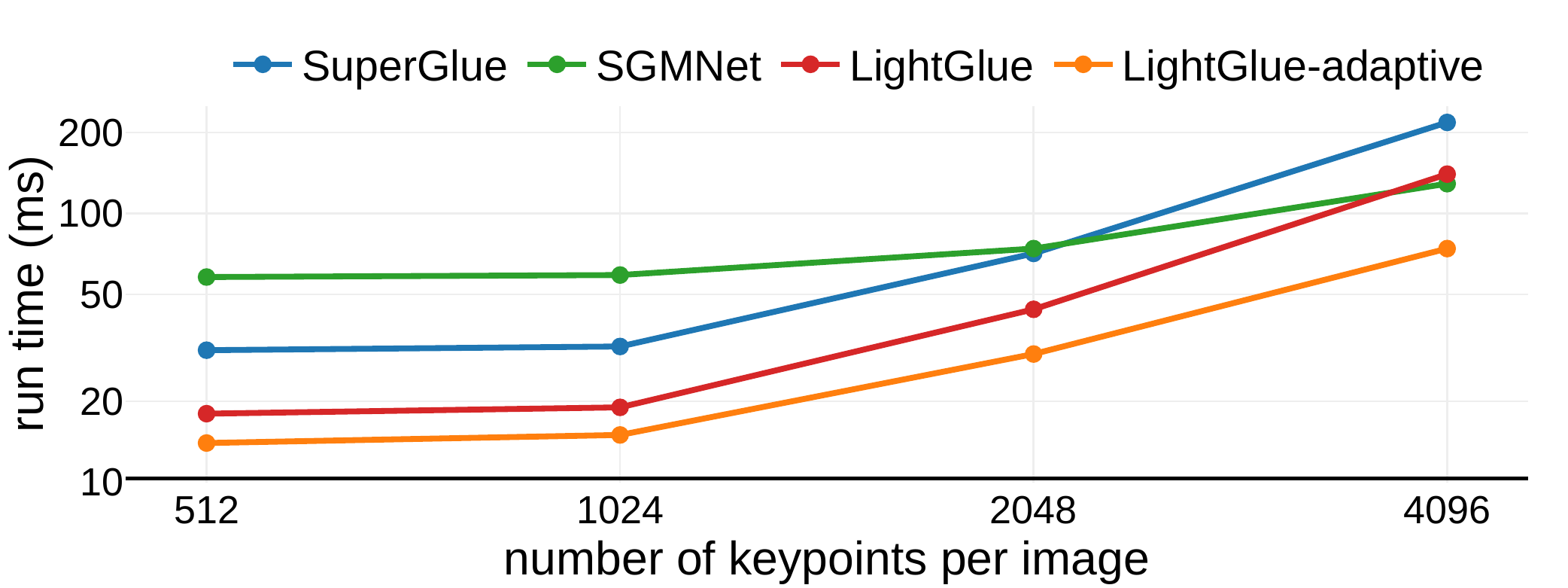}
    \caption{\textbf{Run time vs number of keypoints.}
    The full LightGlue model is 35\% faster than SuperGlue and the adaptive depth and width make it even faster.
    SGMNet is comparably fast only for 4k keypoints and above but is much slower for standard input sizes.
    }%
    \label{fig:timings-keypoints}%
\end{figure}

\PAR{Adaptivity:} By predicting matchability scores and confidences, we can adaptively reduce the computations during a forward-pass on a case-by-case basis.
Table~\ref{tab:adaptive} studies the effectiveness of the two pruning mechanisms -- adaptive depth and width -- on MegaDepth image pairs for different ranges of visual overlap.
For easy samples, such as the successive frames of a video,
the network quickly converges and exits after a few layers, resulting in a 1.86$\times$ speedup.
In cases of low visual overlap, e.g. loop closure, the network requires more layers to converge.
It however rejects confident and unmatchable points early and leaves them out of the inputs to subsequent layers, thus avoiding unnecessary computations.

\PAR{Efficiency:}
Figure~\ref{fig:timings-keypoints} shows run times for different numbers of input keypoints.
For up to 2K keypoints per image, which is a common setting for visual localization,
LightGlue is faster than both SuperGlue~\cite{sarlin2020superglue} and SGMNet~\cite{sgmnet}. 
Adaptive pruning further reduces the run time for any input size.

\section{Conclusion}
This paper introduces LightGlue, a deep neural network trained to match sparse local features across images.
Building on the success of SuperGlue, we combine the power of attention mechanisms with insights about the matching problem and with recent innovations in Transformer.
We give this model the ability to introspect the confidence of its own predictions.
This yields an elegant scheme that adapts the amount of computation to the difficulty of each image pair.
Both its depth and width are adaptive: 1)~the inference can stop at an early layer if all predictions are ready, and 2)~points that are deemed not matchable are discarded early from further steps.
The resulting model, LightGlue, is finally faster, more accurate, and easier to train than the long-unrivaled SuperGlue.
In summary, LightGlue is a drop-in replacement with only benefits.
The code will be released publicly for the benefit of the community.\looseness=-1

\begin{figure*}[t]
    \centering
    \input{figures/qualitative}
    \vspace{-1.5mm}
    \caption{\textbf{Visualization of adaptive depth and width.} From top to bottom, we show three easy, medium and difficult image pairs.
    The left column shows how LightGlue reduces its width: it finds out early that some points ($\color{red}\bullet$) are unmatchable (mostly by visual overlap) and discards non-repeatable points in later layers: ${\color{orange}\bullet} \rightarrow {\color{yellow}\bullet}\rightarrow {\color{green}\bullet}$. This is very effective on difficult pairs. LightGlue looks for matches only in the reduced search space ($\color{blue}\bullet$). The matchability scores (middle column, from non-matchable ${\color{red}\bullet}$ to likely matchable ${\color{green}\bullet}$), help find accurate correspondences and are almost binary. On the right we visualize predicted matches as epipolar in- or outliers.
    We report the run time and stopping layer for each pair. On easy samples, LightGlue stops after only 2-3 layers, running with close to 100 FPS.
    }%
    \label{fig:qualitative}%
\end{figure*}

\begin{figure*}[t]
    \centering
    \input{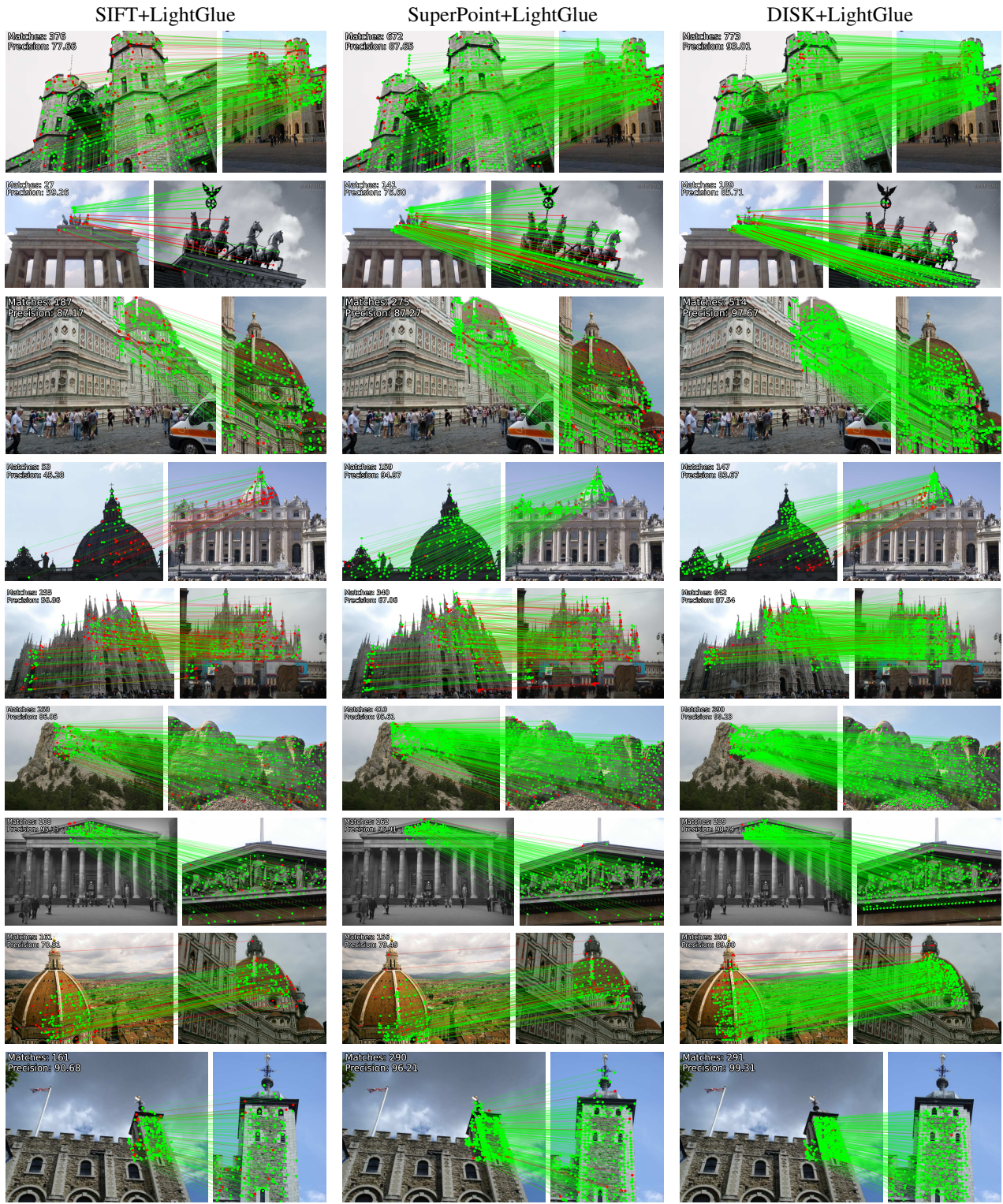}
    \caption{\textbf{Comparison of features produced by LightGlue for different local features.} We compare the outputs of SIFT+LightGlue (left), SuperPoint+LightGlue (middle) and DISK+LightGlue (right).
    }%
    \vspace{5mm}
    \label{fig:local-feats}%
\end{figure*}

\PAR{Acknowledgments:}
We thank Mihai Dusmanu, Rémi Pautrat, and Shaohui Liu for their helpful feedback.

\appendix

\ifproceedings
\pagestyle{plain}
\begin{center}
    {\Large \bf LightGlue:\\Local Feature Matching at Light Speed\par}
    \vspace{0.5cm}
    {
        \large
        Anonymous ICCV submission\\
        Paper ID 4434
    }
\end{center}
\iccvrulercount 0\relax
\setcounter{page}{1}
\fi

\ifproceedings
\section*{Supplementary Material}
In the following pages, we present additional details on the experiments conducted in the main paper.
\else
\section*{\supp}
\fi

\section{Image Matching Challenge}
In this section, we present results obtained on the PhotoTourism dataset of the Image Matching Challenge 2020 (IMC)~\cite{imwchallenge2020} in both stereo and multi-view tracks. 
The data is very similar to the MegaDepth~\cite{li2018megadepth} evaluation, exhibits similar statistics but different scenes.
We follow the standardized matching pipeline of IMC with the setup and hyperparameters of SuperGlue~\cite{sarlin2020superglue}. We run the evaluation on the 3 validation scenes from the PhotoTourism dataset with LightGlue trained with two kinds of local features.

\PAR{SuperPoint:}
For SuperPoint+SuperGlue and SuperPoint+LightGlue, we extract a maximum of 2048 keypoints and use DEGENSAC~\cite{chum2003locally, Chum2005, Mishkin2015MODS} with a threshold on the detection confidence of 1.1 in the stereo track (as suggested by SuperGlue). We do not perform any parameter tuning and reuse our model from the outdoor experiments with adaptive depth- and width, and use efficient self-attention~\cite{flashattention} and mixed-precision during evaluation.

\PAR{DISK:}
We also train LightGlue with DISK local features~\cite{tyszkiewicz2020disk}, a previous winner of the Image Matching Challenge. 
We follow the same training setup as for SuperPoint.
For evaluation, we follow the guidelines from the authors for the restricted keypoint scenario (max 2048 features per image) and use mutual nearest neighbor matching with a ratio test of 0.95 as a baseline. We again use DEGENSAC for relative pose estimation with a threshold of 0.75.

\PAR{Results:}
Table~\ref{tab:imc} reports the evaluation results. We also report the average matching speed over all 3 validation scenes. LightGlue is competitive with SuperGlue both in the stereo and multi-view track, while running 2.5$\times$ faster. Most of these run time improvements are due to the adaptive-depth, which largely reduces the run time for easy image pairs.

LightGlue trained with DISK~\cite{tyszkiewicz2020disk} largely outperforms both the nearest-neighbor matching baseline with ratio test but also SuperPoint+LightGlue. On the smaller thresholds, DISK+LightGlue achieves +8\%/+5\% AUC in the stereo and multi-view tasks compared to our SuperPoint equivalent. With DISK, our model predicts 30\% more matches than SP+LightGlue with an even higher epipolar precision.

\begin{table}[t]
\centering
\footnotesize{
\setlength\tabcolsep{2.7pt}\
\begin{tabular}{lccccccc}
    \toprule
    \multirow{3}{1.9cm}[-.3em]{SfM features (2048 keypoints)}
    & \multicolumn{2}{c}{Task 1: Stereo} & \multicolumn{3}{c}{Task 2: Multiview}
    &\multirow{3}{1.0cm}[-.4em]{Pairs per second}\\
    
    \cmidrule(lr){2-3}
    \cmidrule(lr){4-6}
    &\multicolumn{2}{c}{AUC@K\degree} & \multicolumn{3}{c}{AUC@5\degree@$N$}\\
    \cmidrule(lr){2-3}
    \cmidrule(lr){4-6}
    & 5\degree & 10\degree & 5 & 10 & 25\\
    \midrule
    SP+SuperGlue & 58.64 & 71.07 & 61.88 & 78.97 & 86.75 & 16.2 \\
    \textbf{SP+LightGlue} & \b{59.03} & \b{71.13} &\b{62.87} & \b{79.36} & \b{86.98} & \b{43.4} \\
    \midrule
    DISK+NN+ratio & 57.76 & 68.73  & 59.91 & 78.95 & 87.54 & \b{196.7} \\
    \textbf{DISK+LightGlue} & \b{67.02} & \b{77.82} &\b{67.91} & \b{80.58} & \b{88.35} & 44.5 \\

    \bottomrule
\end{tabular}}
\caption{\textbf{Structure-from-Motion} with the Image Matching Challenge 2020.
We evaluate the stereo track, at multiple error thresholds, and the multi-view track, for various numbers of images $N$.
LightGlue yields better poses than SuperGlue on the multi-view track and significantly reduces the matching time. In combination with DISK, LightGlue improves over SuperPoint+SuperGlue and DISK+NN+ratio in both tracks by a large margin. 
}%
\label{tab:imc}
\end{table}

\PAR{Image Matching Challenge 2021:}
We evaluate the phototourism subset of the IMC 2021~\cite{imwchallenge2021} benchmark, both in the stereo- and multiview track.
We compare our baseline on SuperPoint~\cite{superpoint} and DISK~\cite{tyszkiewicz2020disk} with their respective baselines in a clean setting and in a restricted keypoint setting (max 2048 detections). Furthermore, we compare our best scoring method on IMC 2020, DISK+LightGlue, with tuned versions of DISK~\cite{tyszkiewicz2020disk}, SuperPoint+SuperGlue~\cite{superpoint,sarlin2020superglue} as well as the SfM implementation of the dense matcher LoFTR~\cite{sun2021loftr}. 
Table~\ref{tab:imw2021} reports the experiment.
LightGlue outperforms all approaches with a fair margin.

\PAR{Image Matching Challenge 2023:}
We compete in the IMC 2023~\cite{imwchallenge2023}, which evaluates end-to-end Structure-from-Motion in terms of camera pose accuracy, averaged over multiple thresholds, with a diverse set of scenes beyond phototourism.
We use the default recontruction pipeline of hloc~\cite{sarlin2019coarse} and retrieve 50 pairs per image using NetVLAD~\cite{arandjelovic2016netvlad}.
We average the results over 3 runs to reduce the impact of randomness in the reconstruction pipeline.
On the public / private leaderboards, respectively, SuperPoint+SuperGlue achieves a score of 36.1 / 43.8 (\%), while \textbf{SuperPoint+LightGlue} reaches \textbf{38.4 / 46.1}, which is a \b{+2.3\%} improvement.

\begin{table}[t]
\centering
\footnotesize{\setlength\tabcolsep{2.2pt}\
\begin{tabular}{lccc}
    \toprule
    \multirow{2}{1.9cm}[-.3em]{features + matcher}
    & Task 1: Stereo & Task 2: Multiview & Average \\
    \cmidrule(lr){2-2}
    \cmidrule(lr){3-3}
    \cmidrule(lr){4-4}
    & AUC 5\degree / 10\degree & AUC 5\degree / 10\degree & AUC 5\degree / 10\degree \\
    \midrule
    SP+SGMNet & 29.6 / 43.0 & 60.2 / 71.6 & 44.9 / 57.3 \\
    SP+SuperGlue & 36.5 / 50.5 & 63.3 / 73.8 & 49.9 / 62.2 \\
    SP+\b{LightGlue} & \b{36.7} / \b{50.7} & \b{63.6} / \b{74.4} & \b{50.2} / \b{62.6} \\
    \midrule
    DISK+NN+ratio & 36.3 / 48.5 & 61.5 / 71.6 & 48.9 / 60.1 \\
    DISK+\b{LightGlue} & \b{43.1} / \b{56.6} & \b{66.2} / \b{76.2} & \b{54.7} / \b{66.4} \\
    \midrule
    DISK (8K) +NN+ratio* & 44.6 / 56.2 & 65.0 / 74.4 & 54.8 / 65.3 \\
    SP+SuperGlue* & 44.6 / 58.6 & 66.8 / 77.1 & 55.7 / 67.9 \\
    LoFTR-SfM & 48.4 / 60.9 & 66.4 / 76.1 & 57.4 / 68.5 \\
    DISK (8K)+\textbf{LightGlue} & \textbf{48.7} / \textbf{61.8} & \textbf{68.9} / \textbf{78.2} & \textbf{58.8} / \textbf{70.0} \\

    \bottomrule
\end{tabular}}
\caption{\b{IMC 2021 -- Phototourism.}
*DISK+NN and SP+SG use test-time augmentation while LightGlue does not. To compete with these tuned baselines, we just increase the number of keypoints, e.g. DISK (8K). 
LoFTR-SfM clusters dense matches with SuperPoint detections.  LightGlue outperforms other sparse baselines both in the stereo and multiview task, and even surpasses tuned baselines from the public leaderboard by a large margin.
}%
\label{tab:imw2021}
\end{table}

\section{Additional results}

\PAR{Relative pose estimation:}

Results reported in Section~\ref{sec:rel-pose-outdoor} were computed with a subset of the MegaDepth dataset~\cite{li2018megadepth} as introduced by previous works~\cite{aspanformer, sun2021loftr, matchformer}. However, the images therein overlap with the training set of SuperGlue~\cite{sarlin2020superglue}, the state-of-the-art sparse feature matcher and thus our main competitor.

For a more fair evaluation, we perform an extensive outdoor experiment on the test scenes of our MegaDepth~\cite{li2018megadepth} split, which covers 4 unique phototourism landmarks that SuperGlue was not trained with: Sagrada Familia, Lincoln Memorial Statue,
London Castle, and the British Museum. To balance the difficulty of image pairs, we bin pairs into three categories based on their visual overlap score~\cite{dusmanu2019d2,sarlin2020superglue}, with intervals $[10,30]$\%, $[30,50]$\%, and $[50,70]$\%.
We sample 150 image pairs per bin per scene, totaling 1800 image pairs.
We carefully rerun the experiment with the same setup that was used in Table~\ref{tab:megadepth}.
We report the precision as the ratio of matches with an epipolar error below 3px.
With SIFT~\cite{lowe2004distinctive}, we evaluate the ratio test and SGMNet~\cite{sgmnet} only, as
the original SuperGlue model is not publicly available.

\begin{table}[t]
\centering
\footnotesize{\setlength\tabcolsep{4.1pt}%
\begin{tabular}{rlcccccc}
    \toprule
    \multicolumn{2}{c}{\multirow{2}{*}[-.4em]{features + matcher}}
    &\multirow{2}{*}[-.4em]{\#matches}
    &\multirow{2}{*}[-.4em]{P}
    &\multicolumn{3}{c}{pose estimation AUC}
    &\multirow{2}{0.5cm}[-.4em]{time (ms)}
    \\
    \cmidrule(lr){5-7}
    &&&& @5\degree & @10\degree & @20\degree & \\
    \midrule
    \multirow{3}{*}{\begin{sideways}dense\end{sideways}}
    & LoFTR        & 2231 & 89.8 & 66.4 & 79.1 & 87.6 & 181 \\
    & MatchFormer  & 2416 & 91.2 & 65.2 & 78.1 & 87.4 & 388 \\
    & ASPanFormer  & 4299 & 94.7 & 68.0 & 80.4 & 88.7 & 239 \\
    \midrule[0.2pt]
    \multirow{3}{*}{\begin{sideways}SIFT\end{sideways}}
    & NN+ratio      & 160 & 82.3 & 48.3 & 62.2 & 73.2 & 5.7 \\
    & SGMNet        & 405 & 82.5 & 50.7 & 66.6 & 76.5 & 71.7 \\
    & \b{LightGlue} & 383 & 84.1 & 57.0 & 71.3 & 81.8 & 44.3 \\
    \midrule[0.2pt]
    \multirow{4}{*}{\begin{sideways}SuperPoint\end{sideways}}
    & NN+mutual     & 697 & 49.4 & 37.7 & 50.9 & 62.3 & 5.6 \\
    & SuperGlue     & 712 & 93.0 & 64.8 & 77.5 & 86.6 & 70.0 \\
    & SGMNet        & \b{725} & 89.8 & 61.7 & 74.3 & 83.4 & 74.0 \\
    & \b{LightGlue} & 709 & \b{94.5} & \b{65.5} & \b{77.8} & \b{86.9} & 44.2 \\

    \bottomrule
\end{tabular}}
\caption{\textbf{Relative pose estimation on Megadepth-1800.}
This split is different from Table~\ref{tab:megadepth}.
In contrast to the split used by previous works~\cite{li2018megadepth,sun2021loftr}, this set of test images avoids training overlap with SuperGlue~\cite{sarlin2020superglue}.
LightGlue predicts a similar amount of correspondences but with higher precision (P), pose accuracy (AUC), and
speed than existing sparse matchers. It is competitive with dense
matchers for a fraction of the inference time.
}%
\label{tab:megadepth1800}
\end{table}

Table~\ref{tab:megadepth1800} confirms that LightGlue predicts more accurate correspondences than existing sparse matchers, at a fraction of the time.
Detector-free feature matchers like LoFTR remain state-of-the-art on this task, although by a mere 2\% AUC@5° with LO-RANSAC.

\PAR{Outdoor visual localization:}
For completeness, we also report results on the Aachen v1.1 dataset~\cite{sattler2018benchmarking} and compare our method to recent sparse and dense baselines.
Table~\ref{tab:aachenv1_1} shows that all methods perform similarly on this dataset, which is largely saturated, with insignificant variations in the results.
LightGlue is however far faster than all approaches.

\begin{table}[t]
\centering
\footnotesize{\setlength\tabcolsep{4.7pt}%
\begin{tabular}{lccccc}
    \toprule
    \multirow{2}{1.3cm}[-.4em]{features + matcher}
    & Day & & Night
    & \multirow{2}{1.0cm}[-.4em]{\centering pairs per second}\\
    \cmidrule(lr){2-4}
    & \multicolumn{3}{c}{(0.25m,2°) / (0.5m,5°) / (1.0m,10°)}\\
    \midrule
    LoFTR & 88.7 / 95.6 / 99.0 & & \b{78.5} / 90.6 / 99.0 & - \\
    ASpanFormer & 89.4 / 95.6 / 99.0 & & 77.5 / \b{91.6} / 99.5 & - \\
    \midrule
    SP+SuperGlue & 89.8 / \b{96.1} / \b{99.4} & & 77.0 / 90.6 / \b{100} & \06.4 \\
SP+\b{LightGlue} & \b{90.2} / 96.0 / \b{99.4} & & 77.0 / 91.1 / \b{100} & \b{17.3}\\
    \bottomrule
\end{tabular}}
\caption{\textbf{Outdoor visual localization on Aachen v1.1.} LightGlue achieves similar accuracy with higher throughput.}  %
\label{tab:aachenv1_1}
\end{table}

\begin{figure}[t]
    \centering
    \includegraphics[width=\linewidth]{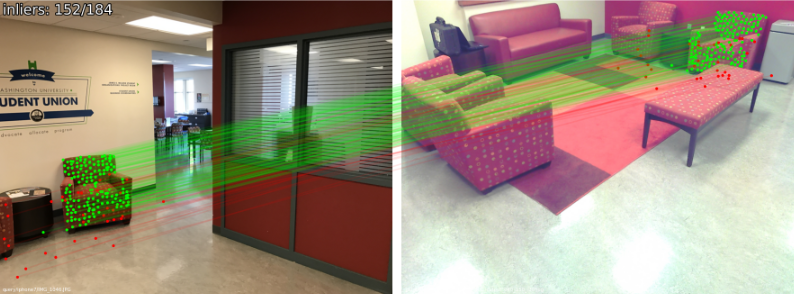}
    \includegraphics[width=\linewidth]{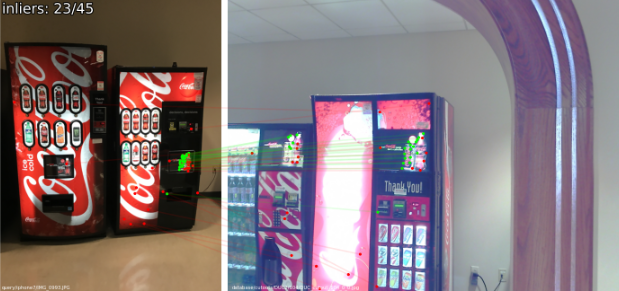}
    \caption{\textbf{Failure cases on InLoc~\cite{taira2018inloc}.} LightGlue sometimes matches repeated objects in the scene with strong texture, instead of the geometric structure.
    }%
    \label{fig:inloc_failure}%
\end{figure}

\PAR{Indoor visual localization:}
We report results for InLoc in Table~\ref{tab:inloc}.
We use hloc and run SuperGlue again for fairness.
For LoFTR and ASpanFormer, report existing results as no code is available.
LightGlue is competitive with SuperGlue and more accurate at (0.25m,10°). 
Differences of $<$2\% are insignificant because each split only has 205/151 queries (1.5\% of difference $\equiv$ 3 queries).
Failures of LightGlue over SuperGlue (6/356 images @1m) are due to more matches on repeated objects (like trash cans), \ie to better matching and weak retrieval -- we show an example in Figure~\ref{fig:inloc_failure}.

\begin{table}[t]
\centering
\footnotesize{\setlength\tabcolsep{4.7pt}%
\begin{tabular}{lccc}
    \toprule
    \multirow{2}{1.3cm}[-.4em]{features + matcher}
    & DUC1 & & DUC2\\
    \cmidrule(lr){2-4}
    & \multicolumn{3}{c}{(0.25m,10°) / (0.5m,10°) / (1.0m,10°)}\\
    \midrule
    LoFTR & 47.5 / 72.2 / 84.8 & & 54.2 / 74.8 / \b{85.5} \\
    MatchFormer & 46.5 / 73.2 / 85.9 & & \b{55.7} / 71.8 / 81.7 \\
    ASpanFormer & \b{51.5} / \b{73.7} / \b{86.4} & & 55.0 / 74.0 / 81.7 \\
    \midrule
    SP+SuperGlue & 47.0 / 69.2 / 79.8 & & 53.4 / \b{77.1} / 80.9 \\
    SP+\b{LightGlue} & 49.0 / 68.2 / 79.3 & & 55.0 / 74.8 / 79.4 \\
    \bottomrule
\end{tabular}}
\caption{\textbf{Indoor visual localization on InLoc.}
LightGlue performs similarly to SuperGlue (within the variability of the dataset).
}%
\label{tab:inloc}
\end{table}

\section{Implementation details}
\label{section:supp:implementation}
\subsection{Architecture}

\PAR{Positional Encoding.}
2D image coordinates are normalized to a range [-1, 1] while retaining the image aspect ratio.
We then project 2D coordinates into frequencies with a linear projection $\*W_p \in \mathbb{R}^{2d/2h}$, where $h$ is the number of attention heads.
We cache the result for all layers.
We follow the efficient scheme of Roformer~\cite{roformer} to apply the rotations to query and key embeddings during self-attention, avoiding quadratic complexity to compute relative positional bias.
We do not apply any positional encoding during cross-attention, but let the network learn spatial patterns by aggregating context within each image.

\PAR{Graph Neural Network:} 
The graph neural network consists of 9 transformer layers with both a self- and cross-attention unit.
The update MLP (Eq.~\ref{eq:update}) has a single hidden layer of dimension $d_h=2d$ followed by LayerNorm, GeLU activation and a linear projection $(2d, d)$ with bias.

Each attention unit has three projection matrices for query, key and value, plus an additional linear projection that merges the multi-head output.
In bidirectional cross attention, the projections for query and key are shared.
In practice we use an efficient self-attention~\cite{flashattention} which optimizes IO complexity of the attention aggregation.
This could also be extended for bidirectional cross attention.
While training we use gradient checkpointing to significantly reduce the required VRAM.

\PAR{Correspondences:}
The linear layers (Eq.~\ref{eq:assignment}) map from $d$ to $d$ and are not shared across layers.
For all experiments we use the mutual check and a filter threshold $\tau = 0.1$.

\PAR{Confidence classifier:}
The classifier predicts the confidence with a linear layer followed by a sigmoid activation.
Confidences are predicted for each keypoint and only at layers $1,..,L-1$, since, by definition, the confidences of the final layer $L$ are 1.
Each prediction is supervised with a binary cross-entropy loss and its gradients are not propagated into the states to avoid impacting the matching accuracy.
The state already encodes sufficient information since it is also supervised for matchability prediction.

\PAR{Exit criterion and point pruning:}
During training we observed that the confidence predictions are less accurate in earlier layers.
We therefore exponentially decay the confidence threshold:
\begin{equation}
    \lambda_l=0.8+0.1 e^{-4\ell/L}\enspace.
\end{equation}
A state is deemed confident if $c_i^\ell>\lambda_\ell$. During inference, we halt the network if $\alpha{=}95\%$ of states are deemed confident.

For point pruning, a point is deemed unmatchable when its predicted confidence is high and its matchability is low:
\begin{equation}
\operatorname{unmatchable}(i) = c_i^l>\lambda_\ell\ \ \& \ \ \sigma_i^\ell < \beta
\label{eq:unmatchable}
\end{equation}

We report an ablation on the exit confidence $\alpha$ in Table~\ref{tab:layerteaser} for relative pose estimation on MegaDepth. Lowering $\alpha$ to 80\% reduces the inference time by almost 50\% compared to our full model, while maintaining competitive accuracy compared to SuperGlue on this task. Reducing the confidence threshold is far more effective in terms of run time - accuracy tradeoff than trimming the model to fewer layers. Stopping the network early mainly sacrifices precision. For our experiments we chose 95\% confidence, which yields on average 25\% run time reduction with hardly any loss of accuracy on downstream tasks. 

\begin{table}[t]
\centering
\footnotesize{\setlength\tabcolsep{4.1pt}%
\begin{tabular}{lcccccc}
    \toprule
    \multirow{2}{*}[-.4em]{Method}
    &\multirow{2}{*}[-.4em]{\#matches}
    &\multirow{2}{*}[-.4em]{P}
    &\multicolumn{3}{c}{pose estimation AUC}
    &\multirow{2}{0.5cm}[-.4em]{time (\%)}
    \\
    \cmidrule(lr){4-6}
    &&& @5\degree & @10\degree & @20\degree & \\
    \midrule
    \b{SP+LightGlue} & \b{613} & \b{96.2} & \b{66.7} & \b{79.3} & \b{87.9} & 100.0 \\
    \cornerarrow layer 7/9 & 705 & 96.0 & 66.2 & 79.1 & 88.0 & 82.4  \\
    \cornerarrow layer 5/9 & 702 & 94.5 & 65.0 & 77.8 & 87.0 & 60.0 \\
    \cornerarrow layer 3/9 & 687 & 90.0 & 64.0 & 76.7 & 85.8 & \b{41.9} \\
    \midrule
    \cornerarrow confidence 98\% & \b{610} & \b{96.2} & 66.6 & 79.3 & 88.0 & 80.5 \\
    \cornerarrow \b{confidence 95\%} & 608 & 95.4 & 66.3 & 79.0 & 87.9 & 70.6 \\
    \cornerarrow confidence 90\% & 607 & 94.5 & 65.9 & 78.5 & 87.2 &  61.5 \\
    \cornerarrow confidence 80\% & 605 & 92.6 & 65.2 & 77.8 & 86.7 & 48.4 \\
    \bottomrule
\end{tabular}}
\caption{\textbf{Evaluation of early-stopping on MegaDepth.}
Matches predicted by deeper layers are more accurate but require more computations with a higher inference time.
Modeling confidences adaptively selects the model depth that yields a sufficient accuracy.
A more conservative stopping, with a higher threshold $\alpha$, yields a higher accuracy at the cost of higher inference time.
$\alpha{=}$95\% yields the best trade-off.
}%
\label{tab:layerteaser}
\end{table}

Here, $\beta=0.01$ is a threshold on how matchable a point is. If Eq.~\ref{eq:unmatchable} holds, we exclude the point from context aggregation in the following layers. This adds an overhead of gather and scatter per layer, but pruning becomes increasingly effective with more keypoints.

\begin{figure}[t]
    \centering
    \includegraphics[width=\linewidth,height=0.4\linewidth]{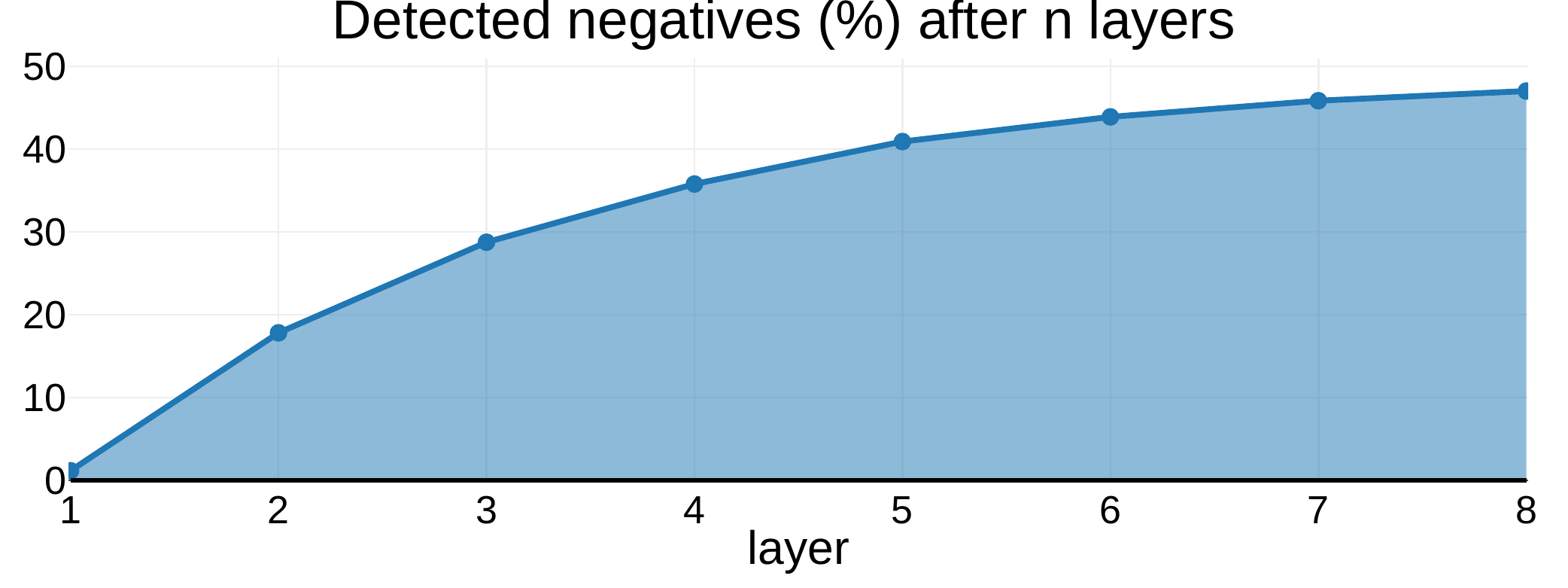}
    \caption{\textbf{Continuous detection of unmatchable points.}
    After just a few layers the network detects many points which are unmatchable, and we exclude them from context aggregation.
    }%
    \label{fig:pruning-layer}%
\end{figure}

In Figure~\ref{fig:pruning-layer} we report the fraction of keypoints excluded in each layer. After just a few layers of context aggregation, LightGlue is confident to exclude $>30\%$ of keypoints early on. Since the number of keypoints have a quadratic impact on run time, as shown in Fig.~\ref{fig:timings-keypoints}, this can largely reduce the number of computations in a forward pass and thus significantly reduce inference time.

\subsection{Local features}

We train LightGlue with three popular local feature detectors and descriptors: SuperPoint~\cite{superpoint}, SIFT~\cite{lowe2004distinctive} and DISK~\cite{tyszkiewicz2020disk}. During training and evaluation, we discard the detection threshold for all methods and use the top-k keypoints according to the detection score. During training, if there are less than k detections available, we append random detections and descriptors.
For SIFT~\cite{lowe2004distinctive} and DISK~\cite{tyszkiewicz2020disk}, we add a linear layer to project descriptors to $d{=}256$ before feeding them to the Transformer backbone.

\PAR{SuperPoint:} SuperPoint is a popular feature detector which produces highly repeatable points located at distinctive regions. We use the official, open-sourced version of SuperPoint from MagicLeap~\cite{superpoint}. The detections are pixel-accurate, i.e. the keypoint localization accuracy depends on the image resolution.

\PAR{SIFT:} We use the excellent implementation of SIFT from vlfeat~\cite{vedaldi10vlfeat} when training on MegaDepth, and SIFTGPU from COLMAP~\cite{schoenberger2016sfm} for fast feature extraction when pre-training on homographies. We observed that these implementations are largely equivalent during training and can be exchanged freely. Also, SIFT features from OpenCV can be used without retraining. 
Orientation and scale are not used in positional encoding.

\PAR{DISK:} 
DISK learns detection and description with a reinforcement learning objective.
Its descriptors are more powerful than SIFT and SuperPoint and its detections are more repeatable, especially under large viewpoint and illumination changes.

\subsection{Homography pre-training}

Following Sarlin \etal~\cite{sarlin2020superglue}, we first pre-train LightGlue on synthetic homographies of real-images.

\PAR{Dataset:}
We use 170k images from the Oxford-Paris 1M distractors dataset~\cite{radenovic2018revisiting}, and split them into 150k/10k/10k images for training/validation/test. 

\PAR{Homography sampling:}
We generate homographies by randomly sampling four image corners. We split the image into four quarters, and sample a random point in each quarter. To avoid degenerates, we enforce that the enclosed area is convex. After, we apply random rotations and translations to the corners s.t. the corners remain inside the image. With this process, we can generate extreme perspective changes while avoiding border artifacts.
This process is repeated twice, resulting in two largely skewed homographies. In interpolation, we then enforce the extracted images to be of size 640x480. 

\PAR{Photometric augmentation:}
The color images are then forwarded through a sequence of strong photometric augmentations, including blur, hue, saturation, sharpness, illumination, gamma and noise. Furthermore, we add random additive shades into the image to simulate occlusions and non-uniform illumination changes.

\begin{figure}[t]
    \centering
    \includegraphics[width=\linewidth,height=0.4\linewidth]{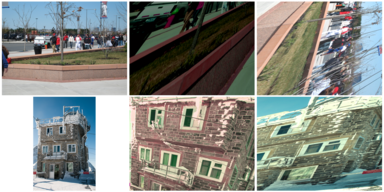}
    \caption{\textbf{Examples of synthetic homographies.} We show the original images (left) and two augmented examples (center and right) resulting from strong perspective transformations and extreme photometric augmentations.
    }%
    \label{fig:homographies}%
\end{figure}

\PAR{Supervision:}
Correspondences with 3px symmetric reprojection error are deemed inliers, and points without any correspondence under this threshold are outliers.

\PAR{Training details:}
We extract 512/1024/1024 keypoints for SuperPoint/SIFT/DISK, and a batch size of 64. The initial learning rate is 0.0001, and we multiply the learning rate by 0.8 each epoch after 20 epochs. We stop the training after 40 epochs (6M image pairs), or 2 days with 2 Nvidia RTX 3090 (for SuperPoint). Our network achieves $>99\%$ recall and $>90\%$ precision on the validation and test set. We also observed that, for fine-tuning, one can stop the pre-training after just one day with only minor losses.

We also experimented with sampling images from MegaDepth~\cite{li2018megadepth} for homography pre-training, and could not observe major differences. Strong photometric augmentations and perspective changes are crucial for training a robust model.

\subsection{Finetuning on MegaDepth}

We fine-tune our model on phototourism images with pseudo ground-truth camera poses and depth images.

\PAR{Dataset:}
We use the MegaDepth dataset~\cite{li2018megadepth}, which contains dense reconstructions of a large variety of popular landmarks all around the globe, obtained through COLMAP+MVS~\cite{schoenberger2016sfm,schoenberger2016mvs}. Following Sun \etal~\cite{sun2021loftr}, we bin each pair by its covisibility score~\cite{dusmanu2019d2}, into ranges $[0.1, 0.3]$, $[0.3, 0.5]$ and $[0.5, 0.7]$. Scenes which are part of the validation and test set in the image matching challenge~\cite{imwchallenge2020} are also excluded from training, resulting in 368/5/24 scenes for training/validation/test. At the beginning of each epoch, we sample 100 image pairs per scene.

Images are resized s.t. their larger edge is of size 1024, and zero-pad images to 1024${\times}$1024 resolution.

\PAR{Supervision:}
Following SuperGlue~\cite{sarlin2020superglue}, we reproject points using camera poses and depth to the other image. Correspondences with a maximum reprojection error of 3 pixels and which are mutually closest are labelled as inliers. A point where the closest correspondence has a reprojection error larger than 5px are is labelled as outlier. Furthermore, we also declare points without depth and no correspondence with a Sampson Error smaller than 3 px outliers.

\PAR{Training details:}
Weights are initialized from the pre-trained model on homographies, Training starts with a learning rate of 1e-5 and we exponentially decay it by 0.95 in each epoch after 10 epochs, and stop training after 50 epochs (2 days on 2 RTX 3090).
The top 2048 keypoints are extracted per image, and we use a batch size of 32. To speed-up training, we cache detections and descriptors per image, requiring around 200 GB of disk space.

\subsection{Homography estimation}
We validate the models capabilities on real homographies on the Hpatches dataset~\cite{balntas2017hpatches}. We follow the setup introduced in LoFTR~\cite{sun2021loftr} and resize images to  a maximum edge length of 480. 

For SuperPoint we extract the top 1024 keypoints with the highest detection score, and report precision (fraction of matches within 3px homography error) and recall (fraction of recovered mutual nearest-neighbour matches within 3px homography error). For LoFTR we only report epipolar precision. Furthermore, we evaluate the models in the downstream task of homography matrix estimation. Following SuperGlue~\cite{sarlin2020superglue}, we report pose estimation results from robust estimation using RANSAC/MAGSAC~\cite{barath2019magsac} and the least squares solution with the weighted DLT algorithm. We evaluate the accuracy of estimated homography by their mean absolute corner distance towards the ground-truth homography.

We use OpenCV with USAC\_MAGSAC for robust homography estimation, and tune the threshold for each method separately. Our reasoning behind this decision, which is in contrast to previous works in feature matching~\cite{sarlin2020superglue, sun2021loftr} which fix the RANSAC parameters, is that we mainly use RANSAC as a tool to evaluate the low-level matches on a downstream task, and we want to minimize the variations introduced by its hyperparameters in order to obtain fair and representative evaluations. Different matches typically require different RANSAC thresholds, and thus a fixed threshold is suboptimal for comparison. For example on outdoor relative pose estimation, tuning the RANSAC threshold yields +7\% AUC@5\degree on SuperGlue, skewing the reported numbers.

\section{Timings}

All experiments were conducted on a single RTX 3080 with 10GB VRAM. We report the timings of the matching process only, excluding sparse feature extraction (which is linear in the number of images) and robust pose estimation. We report the average over the respective datasets.

\begin{figure}[t]
    \centering
    \includegraphics[width=\linewidth,height=0.3\linewidth]{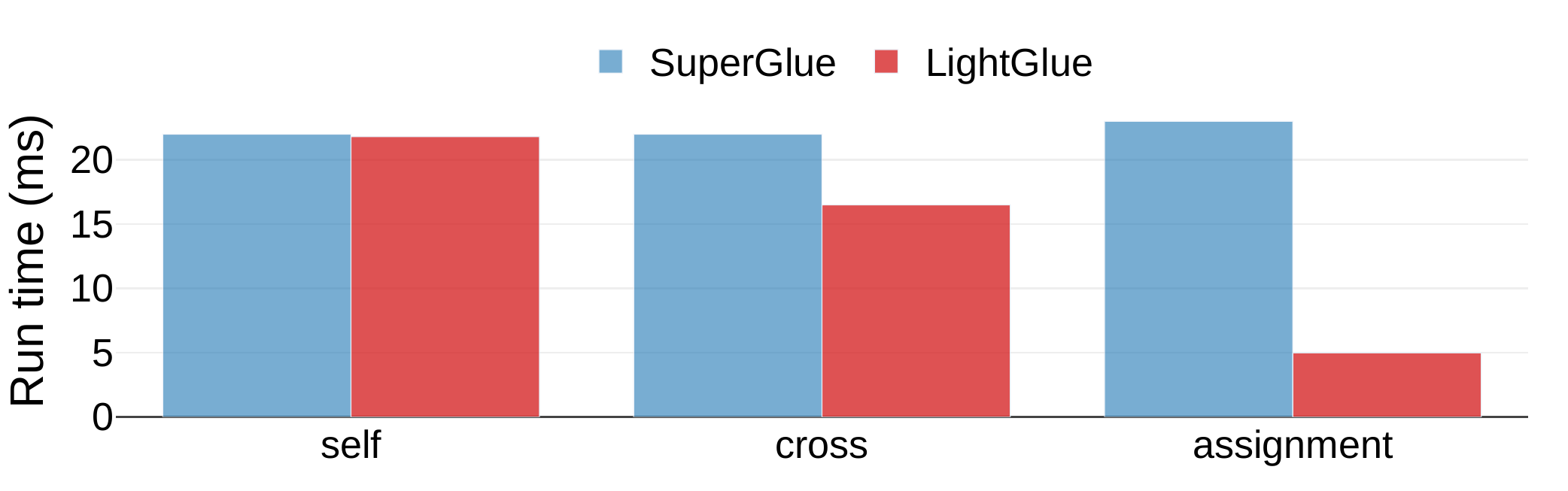}
    \caption{\textbf{Run time breakdown.}
    We evluate the runtime of self-, cross- and partial assignment layers on 1024 keypoints for SuperGlue and LightGlue. Most of LightGlue's default inference time improvements stem from a significantly faster partial assignment layer and reuse of computations in bidirectional cross-attention.
    }%
    \label{fig:timings-steps}%
\end{figure}

In Figure~\ref{fig:timings-steps} we benchmark self-/cross-attention and solving the partial assignment problem against the respective counterparts in SuperGlue~\cite{sarlin2020superglue}. Bidirectional cross-attention reduces the run-time by 33\% by only computing the similarity matrix once. However, the main bottleneck remains computing the softmax over both directions.

Our cheap double-softmax and the unary matchability predictions are significantly faster than solving it using optimal transport~\cite{sinkhorn1967concerning, peyre2019computational}, where 100 iterations are required during training to maintain stability.

In practice, we also use efficient self-attention~\cite{flashattention} and mixed-precision to significantly reduce run time and memory requirements. However, for a fair comparison, we exclude these performance improvements from all experiments except where explicitly stated otherwise.

\section{Qualitative Results}

Figure~\ref{fig:qualitative} shows how LightGlue discards unmatched points and its early stopping mechanism on easy/medium/hard pairs.
Figure~\ref{fig:local-feats} illustrates the matching output for LightGlue with SIFT~\cite{lowe2004distinctive}, SuperPoint~\cite{superpoint} and DISK~\cite{tyszkiewicz2020disk} on some qualitative examples.

{\small
\bibliographystyle{ieee_fullname}
\bibliography{mybib}
}

\end{document}